\documentclass[letterpaper]{article} % DO NOT CHANGE THIS
\usepackage[]{aaai25}  % DO NOT CHANGE THIS
\usepackage{times}  % DO NOT CHANGE THIS
\usepackage{helvet}  % DO NOT CHANGE THIS
\usepackage{courier}  % DO NOT CHANGE THIS
\usepackage[hyphens]{url}  % DO NOT CHANGE THIS
\usepackage{graphicx} % DO NOT CHANGE THIS
\usepackage{natbib}  % DO NOT CHANGE THIS AND DO NOT ADD ANY OPTIONS TO IT
\usepackage{caption} % DO NOT CHANGE THIS AND DO NOT ADD ANY OPTIONS TO IT
\usepackage{algorithm}
\usepackage{amsmath}
\usepackage{amsfonts}
\usepackage{graphicx}
\usepackage{enumitem}
\usepackage{booktabs}
\usepackage{array}
\usepackage{siunitx}  % For aligning numbers at the decimal point
\usepackage{algpseudocode}
\usepackage{multirow}
\usepackage{subcaption}
\newtheorem{definition}{Definition}
\newtheorem{problem}{Problem}

\usepackage{newfloat}
\usepackage{listings}
\DeclareCaptionStyle{ruled}{labelfont=normalfont,labelsep=colon,strut=off} % DO NOT CHANGE THIS
\floatstyle{ruled}
\newfloat{listing}{tb}{lst}{}
\floatname{listing}{Listing}
\title{ST-FiT: Inductive Spatial-Temporal Forecasting
with Limited Training Data}

\author{
    Zhenyu Lei\textsuperscript{\rm 1},
    Yushun Dong\textsuperscript{\rm 2},
    Jundong Li\textsuperscript{\rm 1},
    Chen Chen\textsuperscript{\rm 1}
}
\affiliations{
    \textsuperscript{\rm 1}University of Virginia, 
    \textsuperscript{\rm 2}Florida State University \\
    \{vjd5zr, jundong, zrh6du\}@virginia.edu, yd24f@fsu.edu

}
\usepackage{bibentry}

\begin{document}

\maketitle
\begin{abstract}
Spatial-temporal graphs are widely used in a variety of real-world applications. Spatial-Temporal Graph Neural Networks (STGNNs) have emerged as a powerful tool to extract meaningful insights from this data.
However, in real-world applications, most nodes may not possess any available temporal data during training. For example, the pandemic dynamics of most cities on a geographical graph may not be available due to the asynchronous nature of outbreaks. Such a phenomenon disagrees with the training requirements of most existing spatial-temporal forecasting methods, which jeopardizes their effectiveness and thus blocks broader deployment.
In this paper, we propose to formulate a novel problem of inductive forecasting with limited training data. In particular, given a spatial-temporal graph, we aim to learn a spatial-temporal forecasting model that can be easily generalized onto those nodes without any available temporal training data.
To handle this problem, we propose a principled framework named ST-FiT. ST-FiT consists of two key learning components: temporal data augmentation and spatial graph topology learning.
With such a design, ST-FiT can be used on top of any existing STGNNs to achieve superior performance on the nodes without training data.
Extensive experiments verify the effectiveness of ST-FiT in multiple key perspectives. We present code at: \url{https://github.com/LzyFischer/InductiveST}

\end{abstract}

\section{Introduction}
Spatial-temporal graphs contain both spatial information encoded in graph topology and temporal information encoded in node-associated temporal data~\cite{sahili2023spatio}. In recent years, spatial-temporal graph data has become ubiquitous in a variety of domains such as transportation~\cite{zhang2021traffic}, epidemiology~\cite{wang2022causalgnn}, and social science~\cite{kefalas2018recommendations}.
In these domains, a widely studied task is spatial-temporal forecasting~\cite{zhang2018long,han2021dynamic}, i.e., predicting future temporal dynamics associated with the nodes in given spatial-temporal graphs~\cite{ye2020multi,feng2022adaptive}.
%
% leveraging historical time slots and neighboring time series to predict future data~\cite{geng2019spatiotemporal}. 
% To extract the spatial-temporal dependencies, previous works usually resort to statistical models such as
% Autoregression~(VAR)~\cite{kuethe2011regional} and state space models~\cite{wikle2010general}. Nevertheless, these explorations usually rely on the assumption of linear spatial-temporal dependencies~\cite{guo2021learning}, which thus fall short in capturing complex but common non-linear dependencies in real world data~\cite{lendasse2000non}, such as stock price data with sudden and dramatic fluctuations caused by market sentiment. 
Towards such a goal, Spatial-Temporal Graph Neural Networks~(STGNNs) stand out~\cite{cui2021metro,lan2022dstagnn,bai2020adaptive} due to their capability of synergizing the strengths of Graph Neural Networks~(GNNs) with various sequential forecasting models~\cite{guo2019attention,song2020spatial} to learn the complex spatial-temporal dependencies. As a consequence, STGNNs have been widely adopted in a plethora of real-world applications~\cite{jin2023spatio,zhuang2022uncertainty}.
% This enables STGNNs to exhibit superior forecasting performance compared with other traditional methods~\cite{han2024bigst}.
% treated spatial-temporal data as graphs, where time series are conceptualized as nodes, with their differences~(e.g. geographical distances) represented as edges.

% the pandemic dynamics of most cities on a geographical graph may not be available due to the asynchronous nature of outbreaks

\begin{figure}
    \centering
    \includegraphics[width=\columnwidth]{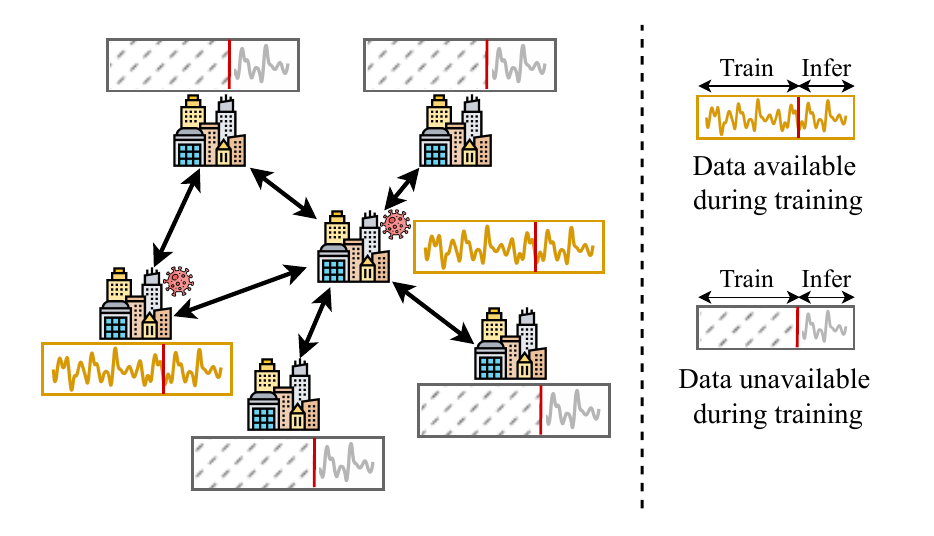}
    % \vspace{-10mm}
    \caption{An exemplary spatial-temporal graph where only the temporal data corresponding to a few nodes is accessible during training: on a geographical graph among different cities, only a few cities have available pandemic dynamics at the current time point (marked in red) due to the asynchronous nature of outbreaks. Here, the virus mark \includegraphics[scale=0.025]{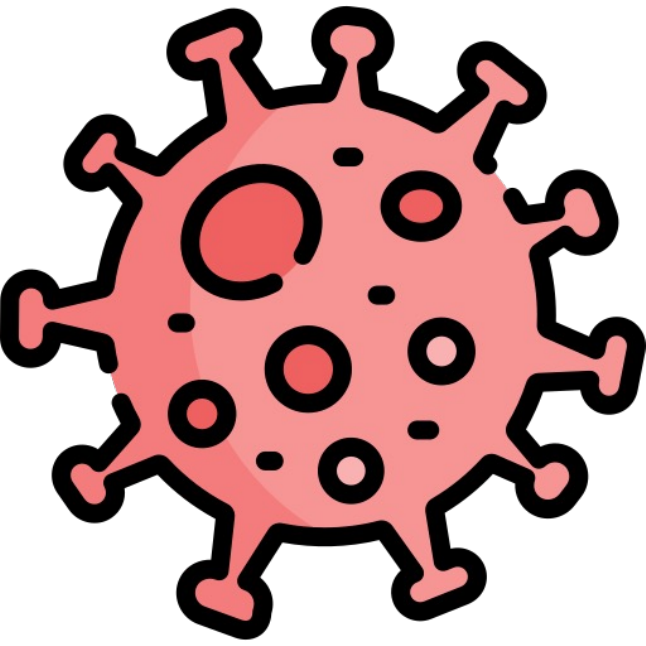} denotes the cities that have gone through outbreaks.}
    \vspace{-5mm}
    \label{fig:teaser_1}
\end{figure}

Despite advancements, most existing STGNNs require that all nodes in the given spatial-temporal graphs should have temporal data (e.g., time series data) during training~\cite{wang2020traffic,li2021spatial}, such that the unique temporal dependency for each node can be easily captured~\cite{shin2024pgcn}. 
With the captured temporal dependencies, the STGNNs could make predictions for each node effectively.
%
% are designed under the assumption that the time series data corresponding to all nodes are available in the training stage. 
However, in real-world scenarios, most nodes may not have available temporal data during training~\cite{gupta2023frigate,wu2022inductive}. We present an exemplary case in Figure~\ref{fig:teaser_1}. Specifically, facing a sudden pandemic such as COVID-19, due to the asynchronous nature of pandemic outbreaks, the pandemic dynamics (e.g., the tendency of confirmed case number) of most cities on a geographical graph may not be available at a given time point (marked in red)~\cite{atchade2022overview,panagopoulos2021transfer}. In such cases, existing STGNNs perform poorly in cities without available temporal training data.
% In such cases, since training requirements are not met, existing STGNNs can hardly achieve satisfying performance in cities without available temporal training data.
%
Therefore, to enable wider deployment, the forecasting model should generalize the learned temporal dependencies to the nodes without any temporal data during training, referred to as \textit{inductive forecasting with limited training data}.
A few recent studies have made early explorations to such a problem~\cite{tang2022domain}. For example, domain adaptation strategies~\cite{fang2022transfer,wang2021spatio} have been adopted to generalize dependencies from nodes with abundant temporal data.
% achieve satisfactory generalization performance by adapting the learned dependencies from nodes with abundant temporal data.
However, these works overwhelmingly focus on the generalization between different spatial-temporal graphs, ignoring granular temporal dependency differences within the same graph. Furthermore, they usually require costly fine-tuning~\cite{zhou2022domain}, which limits their efficiency for real world scenarios~\cite{guo2023self,li2020autost}.
% making them unable to satisfy the need to perform efficient inference in a wide range of real-world scenarios.
%
Therefore, despite the practical significance, the problem of enabling inductive forecasting with limited training data remains underexplored.

It is worth mentioning that inductive forecasting with limited training data on spatial-temporal graphs presents three challenges. (1)~\textbf{Limited Temporal Dependencies.} With the temporal data corresponding to only a limited number of nodes available, the forecasting model can only extract limited temporal dependencies~\cite{lachapelle2024nonparametric,liu2012sparse}. However, nodes without any temporal training data may still require different temporal dependencies to perform accurate forecasting~\cite{wijsen2018temporal}. 
% We refer to such a disagreement in temporal dependencies as \textit{temporal heterogeneity}.
% , which is illustrated in Figure~\ref{fig:teaser2_a}.
%
Therefore, the first challenge is to learn diversified temporal dependencies for better generalization.
% to achieve satisfactory generalization performance onto different nodes in the given spatial-temporal graphs.
%
(2)~\textbf{Diverse Spatial Dependencies.}
The spatial-temporal graph topology encodes spatial dependencies, crucial for generalizing the learned temporal dependencies between neighboring nodes~\cite{liang2022spatial}. However, the topology may exhibit different patterns of spatial dependencies in different local areas~\cite{park2014spatially}. For example, during the pandemic,
% in the geographical graph of the cities under pandemic,
% (as in Figure~\ref{fig:teaser2_b}) 
geographically neighboring cities may exhibit both similar and distinct pandemic dynamics due to varying interactions (e.g., different volumes of population migration~\cite{gibbs2020changing}). Hence, the second challenge is to equip the forecasting model with generalization capability across different spatial dependencies.
(3)~\textbf{\textit{Inference Efficiency.}} Most existing explorations aiming to handle differences in spatial and temporal dependencies require costly fine-tuning processes~\cite{zhou2022domain,ouyang2022causality}, which makes efficient inference difficult in real-world scenarios.
% it difficult to satisfy the need to perform efficient inference in a wide range of real-world scenarios.
Our third challenge is to avoid additional computational costs and achieve efficient inference on nodes with no available temporal data for training.

To address the challenges above, we introduce ST-FiT~(inductive \underline{S}patial-\underline{T}emporal \underline{F}orecasting with l\underline{i}mited \underline{T}raining data), a novel framework that generalizes to different spatial-temporal dependencies without fine-tuning. Specifically, ST-FiT consists of two learning modules for above challenges. To handle the first challenge, ST-FiT introduces a temporal data augmentation module. This module learns the manifold where the available training temporal data lies and generates new temporal data close to it to enrich the training set. 
% why temporal augmentation could enrich dependencies? quantilization of temporal dependencies (frequency).
In this way, the training temporal dependencies can be enriched
% why enriched could enhance the generalization? cover? how? increase the possibility to cover? the true effect is learn a global one.
% converse temporal dependencies
, such that the generalization ability to different temporal dependencies can be enhanced for the forecasting model. 
To handle the second challenge, ST-FiT is equipped with a spatial graph topology learning module. With this module, spatial dependencies (represented as edges in spatial-temporal graphs) between new and existing temporal data can be generated, and existing spatial dependencies can be refined as well.
We formulate an optimization problem with an iterative training strategy for these modules. As such, ST-FiT is enabled to perform inductive forecasting with any STGNN backbone while avoiding costly fine-tuning. Such generalization capability and flexibility significantly broaden a broader range of applicable scenarios compared with other alternative spatial-temporal forecasting models.
Empirical evaluations on three commonly used real-world datasets corroborate the effectiveness of ST-FiT in multiple key perspectives. Our contributions are summarized as follows: 
\begin{itemize}
    \item \textbf{Problem Formulation.} We formulate a novel problem of inductive forecasting with limited training temporal data, which aligns with the setting associated with a wider spectrum of real-world applications.
    \item \textbf{Framework Design.} We design a new framework named ST-FiT to handle the key challenges associated with the problem studied and achieve superior performance compared to other alternatives.
    \item \textbf{Experimental Evaluations.} We conduct comprehensive experiments on three commonly used real-world datasets to verify the effectiveness of our proposed framework.
\end{itemize}

\section{Problem Definition}
\label{sec:2}
In this section, we first present the notations used throughout this paper. Then we introduce two key definitions, including \textit{Spatial-Temporal Graph} and \textit{Spatial-Temporal Forecasting}. Finally, we introduce a novel problem of \textit{Inductive Spatial-Temporal Forecasting with Limited Training Data}.

\begin{figure*}
    \centering
    \vspace{2mm}
    \includegraphics[width=0.98\textwidth]{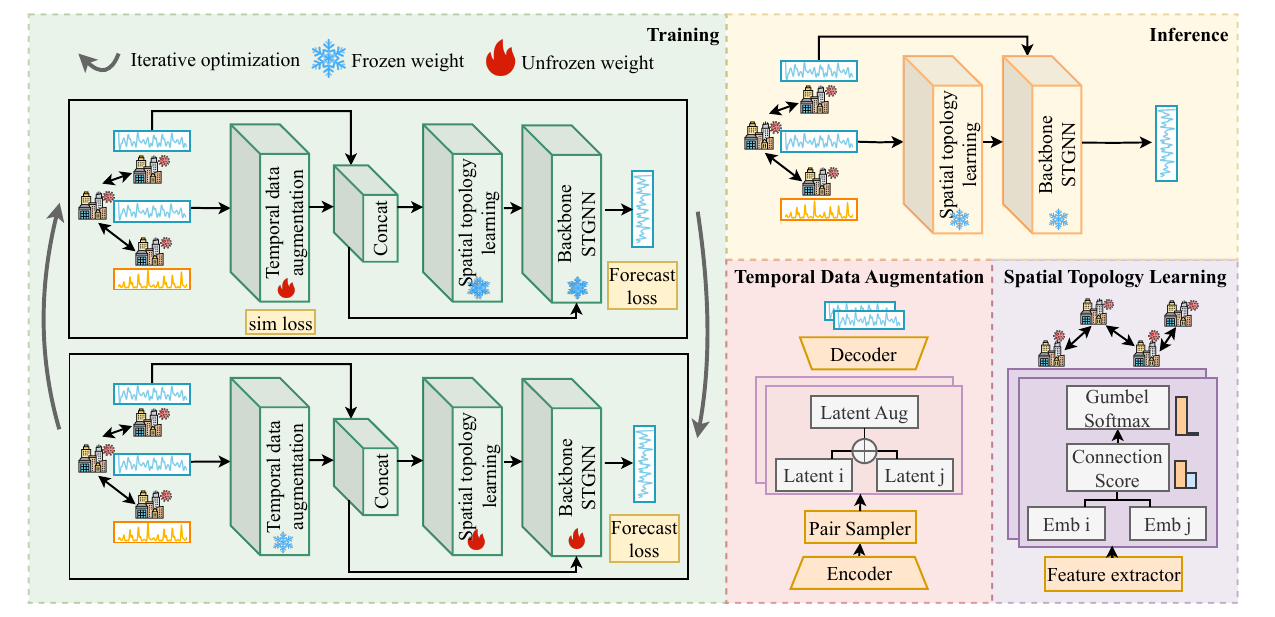}
    \caption{An overview of ST-FiT, including a STGNN backbone, temporal data augmentation, and spatial topology learning.}
    \label{fig:overview}
\end{figure*}

\noindent{\textbf{Notations.}} 
% In this paper, without further specification, We use calligraphic letters (e.g., $\mathcal{D}$), bold uppercase letters~(e.g., $\textbf{D}$), bold lowercase letters~(e.g., $\textbf{d}$), and normal uppercase/lowercase letters~(e.g., $D/d$) to denote sets, matrices, vectors, and scalars, respectively. Besides, for any matrix~(e.g., $\textbf{D}$), we represent its $i$-th row as $\textbf{d}_i$, its ($i$,$j$)-th entry as $\textbf{D}_{ij}$, and its transpose as $\textbf{D}^T$. 
We denote an attributed graph as $\mathcal{G} = (\mathcal{V}, \mathcal{E}, \boldsymbol{A}, \boldsymbol{X})$, where $\mathcal{V}$ is the set of nodes and $N = |\mathcal{V}|$ is the number of nodes. $\boldsymbol{X} \in \mathbb{R}^{N \times C}$ represents the feature matrix of the nodes in $\mathcal{V}$, where $C$ is the dimension number of node features. $\mathcal{E}$ denotes the set of edges, where the edge between node $v_i$ and $v_j$ is denoted as $e_{ij} = (v_i, v_j)$. $\boldsymbol{A} \in \{0, 1\}^{N \times N}$ is the adjacent matrix, where $\boldsymbol{A}_{ij} = 1$ indicates that an edge exits between $v_i$ and $v_j$, otherwise $\boldsymbol{A}_{ij} = 0$.

\begin{definition}
\label{def_1}
\textbf{Spatial-Temporal Graph.} A spatial-temporal graph $\{\mathcal{G}^t\}_{t=1}^{T}$ contains a sequence of graphs $\mathcal{G}^t$ ($1 \leq t \leq T$), where $t$ is the current time step and $T$ is the total number of time steps. Here each $\mathcal{G}^t$ is described as $(\mathcal{V}, \mathcal{E}, \boldsymbol{A}, \boldsymbol{X}^t)$, where $\boldsymbol{X}^t \in \mathbb{R}^{N \times C}$ denotes the node features at time step $t$. The graph topology described by $\boldsymbol{A}$ reveals the spatial dependency, while the temporal data corresponding to each node consists of all node features across all time steps.
%
% $\{\mathcal{G}^t\}_{t=1}^{T}$ $T$ steps. Each $\mathcal{G}^t$ is described as $(\mathcal{V}, \mathcal{E}, \boldsymbol{A}, \boldsymbol{X}^t)$, where $X^t \in \mathbb{R}^{N \times C}$, with $C$ representing the number of features for each node. 
\end{definition}

\begin{problem}
\label{problme1}
    \textbf{Spatial-Temporal Forecasting} Given a sliding window of $\kappa$ time steps in a spatial-temporal graph $\mathcal{G}^{t-\kappa:t}$ , our goal is to learn a function $f$ to predict features $\boldsymbol{X}^{t+1:t+\tau}$ in following $\tau$ time steps. Here $(\cdot)^{t_1: t_2}$ denotes a sequence (ordered by time steps) with a length of $t_2 - t_1 + 1$, where the input at time step $t_i$ ($t_1 \leq t_i \leq t_2$) is placed at the $(t_i - t_1 + 1)$-th position.
\end{problem}

Based on Definition~\ref{def_1} and Problem~\ref{problme1} above, we then present the problem of \textit{Inductive Spatial-Temporal Forecasting with Limited Training Data} below.

\begin{problem}
\label{problem2}
    \textbf{Inductive Spatial-Temporal Forecasting with Limited Training Data.} Given a spatial-temporal graph with $T_{train}$ time steps $\mathbb{G} = \{\mathcal{G}^t\}_{t=1}^{T_{train}}$ where the features of only a small subset of nodes (denoted as $\mathcal{V}_{train}$ and fixed in all steps) are available, we aim to learn a forecasting model $f$ to accurately predict the whole node feature matrix $\boldsymbol{X}^{T_{train}+1:T_{train}+\tau}$ from step $T_{train}+1$ to $T_{train}+\tau$.
% \begin{equation}
%     X^{t+1:\tau} = f(\mathcal{G}^{t-\kappa:t}),
% \end{equation}
% to minimize the following error during test time
% \begin{equation}
%     \mathcal{L}_{test}=\sum_{v \in \mathcal{V}} \sum_{t=T_{test}}^{T-\tau} \sum_{j=1}^{\tau} \mathcal{L}_{e r r}\left(\hat{\boldsymbol{x}}_v^{t+j}, \boldsymbol{x}_v^{t+j}\right),
% \end{equation}
% where $\hat{\boldsymbol{x}}_v^{t+j}$ and $\boldsymbol{x}_v^{t+j}$ are predicted and observed features at node $v$ and time $t+j$ respectively, and $\mathcal{L}_{e r r}$ is an error function such as mean squared error.
\end{problem}

\section{Methodology}
% 1. overview
%     1. component-based, what do I have, augmentation (alternating training, designed loss, mix-up, manifold), graph learning (gumbel-soft, sparse, ), backbone stgnn (why do I need this?)
%     2. alternative training? optimization? objective and optimization?
%     3. masked node loss. 
%     4. during inference? keep only training nodes. 
%     5. do we need to provide the overall training loss L1, L2, L3, L4
% 2. structure
%     1. augmentation, objective, graph learning, objective, backbone, objective,
%     2. alternating: 1. augmentation, 2.structure learning and backbone stgnn
% 3. thinking
%     1. from probabilistic? or pure maximization? 
%         1. vae and gumbel-softmax all need probabilistic
%         先自己写出来

\subsection{Overview}
An overview of ST-FiT is shown in Figure~\ref{fig:overview}. Specifically, ST-FiT consists of 3 modules:
(1) \textit{Temporal Data Augmentation} aims to generate diverse temporal data (in the format of time series) via learning the manifold where the accessible training temporal data lies. (2) \textit{Spatial Topology Learning} aims to generate the spatial dependencies between existing and newly generated time series and refine the spatial dependencies between existing time series. (3) \textit{STGNN Backbone}. ST-FiT is plug-and-play, i.e., any STGNN model can be adopted as the backbone of this framework to achieve forecasting on nodes without training temporal data.
%
% is compatible with any STGNN backbone  to predict future temporal dynamics, with augmented temporal data and learned topology as input. Unless otherwise stated, we instantiate the STGNN backbone as STGCN~\cite{yu2017spatio} due to its widespread adoption and compatibility for inductive forecasting.
To effectively optimize the three learning modules, we formulate the optimization problem first and propose an iterative approach to solve it. We introduce the three modules below.

\subsection{STGNN Backbone}

% Here we briefly introduce the formulation of the STGNN backbone model. 
Specifically, an STGNN backbone model takes the temporal data (encoded with temporal dependencies) and the spatial-temporal graph topology (encoded with spatial dependencies) as input, and outputs predicted temporal data. 
% as the forecasting results. 
In particular, we assume the temporal data is in the format of time series; the input and output time series are with a length of $\kappa$ and $\tau$, respectively. We formulate the STGNN backbone as
\begin{equation}
    \boldsymbol{x}^{t+1:t+\tau} = h(\boldsymbol{x}^{t-\kappa:t}, \boldsymbol{A}),
    \label{eq:forecast}
\end{equation}
where $h$ denotes the mapping given by the STGNN backbone; $\boldsymbol{x}^{i:j}$ denotes the time series given by the node feature matrix $\boldsymbol{X}^t$ from time step $i$ to $j$; $\boldsymbol{x}^{t-\kappa:t}$ and $\boldsymbol{x}^{t+1:t+\tau}$ represent the input and output time series, respectively. 

\subsection{Temporal Data Augmentation}

To handle the challenge of limited temporal dependencies, we resort to data augmentation to enrich the training temporal data.
% such that more diverse temporal dependencies can be involved in training. 
However, commonly used methods such as cropping and adding noise in input or latent space are inadequate since they cannot generate new dependencies.
% most of the existing data augmentations in the time series domain are based on straightforward transformations (e.g., cropping and warping) or adding noise in either input space or latent space. These straightforward approaches cannot generate temporal data with new dependencies, and thus they cannot be directly adopted to handle the challenge above.
% manifold lie in hidden space?
In order to achieve our goal, we propose to learn a manifold where all available temporal data lies in the hidden space. Then, we select data points that lie between the points associated with the accessible training temporal data on the manifold, such that new data following new temporal dependencies can be generated.
% new temporal dependencies ? why our model could generate new dependencies?
We refer to this strategy as temporal data mix-up on the learned temporal manifold.
%
% data augmentation is a very straightforward solution due to its excellent ability to enhance diversity and generalizability~\cite{sandfort2019data}. Most of the existing data augmentations in the time series domain are heuristic based on simple transformations such as cropping and warping, or adding noise in either input spaces or latent spaces. More recent works introduced mix-up techniques~\cite{eldele2023contrastive,darlow2023tsmix} from the image domain, leveraging its power to generate diverse temporal data as convex combinations of two random selected time series during training. However, most of the generated  time series struggle to lie on the manifold of original temporal data~\cite{demirel2024finding}. Nevertheless, recent research from diverse domains shows that such desiderata of generating better data could be satisfied by mix-up on the manifold~\cite{zhou2023improving,el2024augment,verma2019manifold}, which inspired us to conduct mix-up on the manifold where available training temporal data lies.
%
% Specifically, our proposed temporal data augmentation is composed of three core designs: a Variational Autoencoder~(VAE) model to encode sliding windows of temporal data into hidden representations, the mix-up on the manifold, and novel objective to learn the manifold where available data lies. 
Specifically, we first input the time series from each sliding window into a Variational Auto-Encoder (VAE)~\cite{kingma2013auto}, from which we can derive the representations that characterize the manifold of the temporal data in the hidden space. Denote $\xi$ as a time step between $\kappa + 1$ and $T_{train} - \tau$. We achieve the above operations with
\begin{equation}
\begin{aligned}
    \boldsymbol{\mu}_v, \boldsymbol{\sigma}_v &= \operatorname{Encoder}(\boldsymbol{x}^{\xi-\kappa:\xi+\tau}_v) \;\text{and} \\
    \boldsymbol{z}_v &= \operatorname{Sample}(\boldsymbol{\mu}_v, \boldsymbol{\sigma}_v),
\label{eq:aug_1}
\end{aligned}
\end{equation}
where $v$ denotes the node (in the given spatial-temporal graph) we are focusing on; $\boldsymbol{\mu}_v$ and $\boldsymbol{\sigma}_v$ are the mean and standard deviation characterizing the Gaussian distribution to be sampled from the latent space; $\kappa + \tau$ denotes the length of the input time series.
We assume that $\boldsymbol{z}_v$ from all nodes come from a unified manifold in the hidden space.
Then, we can perform temporal data mix-up on the learned manifold. Specifically, we randomly sample $K$ pairs of time series and generate new hidden representations from each pair by
\begin{equation}
\begin{aligned}
    \boldsymbol{\hat{z}}_{v} = \lambda \cdot &\boldsymbol{z}_{v_i} + (1 - \lambda) \cdot \boldsymbol{z}_{v_j},
\label{eq:aug_2}
\end{aligned}
\end{equation}
where $\mathcal{U}$ is the sampled pair set ($\{v_i, v_j\} \in \mathcal{U}, |\mathcal{U}| = K$); $\boldsymbol{\hat{z}}_{v}$ is the generated hidden representation; $\lambda$ is the mix-up ratio. Since the positions for $\boldsymbol{z}_{v_i}$ and $\boldsymbol{z}_{v_j}$ are symmetric, $\lambda$ naturally falls between $0$ and $0.5$. We note that this does not rigorously guarantee that $\boldsymbol{\hat{z}}_{v}$ lies on the learned manifold. However, this generally enables us to generate $\boldsymbol{\hat{z}}_{v}$ close to the learned manifold, which empirically leads to effective utility improvements.
Finally, we transform the generated hidden representations back to input space with a VAE decoder to obtain new temporal data by
\begin{equation}
    \boldsymbol{\hat{x}}^{\xi-\kappa:\xi+\tau}_{v} = \operatorname{Decoder}(\boldsymbol{\hat{z}}_{v}),
\label{eq:aug_3}
\end{equation}
where $\boldsymbol{\hat{x}}^{\xi-\kappa:\xi+\tau}_{v}$ is the generated temporal data.

To optimize the learnable parameters in the encoder and decoder, we design a training objective to maximize the similarity between the available pair $(\boldsymbol{z}_{v_i}, \boldsymbol{z}_{v_j})$ and the newly generated $\boldsymbol{\hat{z}}_{v}$. The rationale is that the newly generated $\boldsymbol{\hat{z}}_{v}$ should preserve similar temporal dependencies from both $\boldsymbol{z}_{v_i}$ and $\boldsymbol{z}_{v_j}$. We formulate the optimization goal as
%
% \begin{equation}
% \begin{aligned}
%     \mathcal{L}_{sim} &= \sum_\xi \sum_{\{v_i, v_j\}\in \mathcal{U}}\lambda\cdot\operatorname{cosine}(\boldsymbol{\hat{x}}^{\xi-\kappa:\xi}_{v_k}, \boldsymbol{x}^{\xi-\kappa:\xi}_{v_i}) \\
%     &+ (1 - \lambda)\cdot\operatorname{cosine}(\boldsymbol{\hat{x}}^{\xi-\kappa:\xi}_{v_k}, \boldsymbol{x}^{\xi-\kappa:\xi}_{v_j}),     
% \end{aligned}
% \end{equation}
%
\begin{equation}
\begin{aligned}
    \mathcal{L}_{sim} &= \sum_\xi \sum_{\{v_i, v_j\}\in \mathcal{U}}\lambda\cdot\operatorname{cosine}(g(\{v_i, v_j\}, \xi, \tau), \boldsymbol{z}_{v_i}) \\
    &+ (1 - \lambda)\cdot\operatorname{cosine}(g(\{v_i, v_j\}, \xi, \tau), \boldsymbol{z}_{v_j}),     
\end{aligned}
\end{equation}
where function $g(\cdot, \cdot, \cdot)$ denotes the function to generate new time series with node pair (the first parameter), time step (the second parameter), 
% whether use \xi - \kappa, \xi + \tau
and the steps that go beyond the given time step (the third parameter) through the encoder-decoder design; function cosine$(\cdot, \cdot)$ takes two vectors as input and outputs their cosine similarity.
To further ensure that the generated temporal data with $g(\{v_i, v_j\}, \xi, \tau)$ reflects a consistent dependency across all $\kappa +\tau$ steps, we propose to utilize the backbone STGNN model to perform forecasting on the generated temporal data. Our rationale is that accurate forecasting with the backbone model reveals the existence of a consistent dependency across all time steps. We propose to formulate such an objective as
\begin{equation}
\begin{aligned}
    \mathcal{L}_{fst} = \sum_\xi \sum_{\{v_i, v_j\}\in \mathcal{U}} \mathcal{L}_{e r r} (h(g&(\{v_i, v_j\}, \xi, \tau)_{[:\xi]}, \boldsymbol{A}), \\ &g(\{v_i, v_j\}, \xi, \tau)_{[-\tau:]}),
\end{aligned}
\end{equation}
where $(\cdot)_{[:\xi]}$ and $(\cdot)_{[-\tau:]}$ denote the first $\xi$ and last $\tau$ steps in a given sequence, respectively; function $h$ denotes the backbone STGNN that takes a sequence with a length of $\xi$ and adjacency matrix and outputs the predicted sequence with a length of $\tau$; $\mathcal{L}_{e r r}$ measures the difference between two input sequences, e.g., the element-wise mean squared error.

\subsection{Spatial Topology Learning}
% % 1. learn a structure is better
% % 2. sparse structure is better
% 1. overfitting heterogeneity
% 2. sparse dynamic diverse learn gumbel soft-max

To handle the challenge of diverse spatial dependencies, we propose to refine existing spatial topology and generate the topology between the generated and existing temporal data. As such, we are able to better adapt the spatial topology to fit the predictive capability of the STGNN backbone.
Intuitively, this module should be agnostic to the number of nodes, such that it meets the need for inductive forecasting. Meanwhile, the learned spatial topology should be naturally discrete and sparse so that it only encodes key patterns of spatial dependencies~\cite{hu2022improving}.
To achieve the goals above, we propose to leverage the Gumbel-Softmax reparameterization to learn a sparse graph topology based on the node features~\cite{shang2021discrete}.

Specifically, we first use a Multi-Layer Perceptron (MLP) encoder to transform each sliding window of time series into a hidden representation. Then, we use another MLP maps the hidden representations for each pair of nodes $v_i, v_j$ to a scalar $\boldsymbol{P}_{ij} \in [0,1]$. We utilize the matrix $\boldsymbol{P}$ to parameterize a Bernoulli distribution between every node pair. By drawing samples from the Bernoulli distributions, we are able to construct a refined adjacency matrix $\boldsymbol{\tilde{A}}$ to characterize the learned spatial topology, i.e., ~$\boldsymbol{\tilde{A}}_{ij} \sim Ber(\boldsymbol{P}_{ij})$. Note that we apply the Gumbel reparameterization trick~\cite{franceschi2019learning} to enable the gradient to flow through $\boldsymbol{\tilde{A}}$, such that gradient-based techniques can be adopted to optimize $\boldsymbol{P}$. We formulate the procedure to derive $\boldsymbol{\tilde{A}}$ as  
\begin{equation}
    \boldsymbol{\tilde{A}}_{i j} = \operatorname{Gumbel-Softmax}(\boldsymbol{P}_{ij}, s),
    \label{eq:topology}
\end{equation}
where $s$ is the temperature parameter of Gumbel-Softmax.
However, based on the formulation given above, it becomes difficult then to impose $l_1$ norm as a regularization to achieve sparse graph topology.
To enforce the learned spatial topology to be sparse, we propose to transform the learned matrix $\boldsymbol{P}_{ij}$ with a threshold~$\epsilon$,~i.e.,
\begin{equation}
    \boldsymbol{\hat{P}}_{ij} = \operatorname{Sigmoid}({\rm e}^{(\boldsymbol{P}_{ij} - \epsilon) / \phi}),
\end{equation}
where $\phi$ is the temperature. Intuitively, $\boldsymbol{P}_{ij} < \epsilon$ will make it less likely to generate an edge between node $v_i$ and $v_j$.

\subsection{Optimization Strategy and Inference}

The training objectives of ST-FiT are three-fold: (1) Generate diverse temporal data that lies close to the learned manifold;
% based on the available temporal data for training
(2) Refine spatial topology based on diverse spatial dependencies; (3) Capture key spatial-temporal dependencies for forecasting with the STGNN backbone. 
% To achieve the objectives above, we propose to optimize three modules simultaneously.
%
% there lies a causal relationship between the temporal data augmentation module and the remaining module. To be specific, generating temporal data lying on the manifold of existing data requires the captured temporal dependencies of STGNN backbone, while learning a generalizable STGNN requires training on time series data with diverse temporal dependencies. Following this intuition
%
However, we note that the optimization of temporal data augmentation and the other two modules are intertwined, since (1) it requires the STGNN backbone to perform prediction ; and (2) it requires the refined spatial dependencies from the spatial topology learning module as the input (as in $\mathcal{L}_{fst}$).
As such, we propose to formulate the overall optimization problem and solve it with an iterative training strategy, i.e., training the temporal data augmentation module and the other two modules iteratively. We refer to the two optimization processes in each iteration as the \textit{Phase 1} (optimizing the temporal data augmentation module) and the \textit{Phase 2} (optimizing other modules), respectively. 

In \textit{Phase 1}, we optimize the temporal data augmentation with gradient-based optimization techniques, while the parameters of other modules are frozen. Formally, we have
\begin{equation}
    \boldsymbol{\theta}^{aug}_{\text{epoch+1}} = \boldsymbol{\theta}^{aug}_{\text{epoch}} - \eta\cdot\nabla_{\boldsymbol{\theta}^{aug}}\mathcal{L}_{aug},
    \label{eq:update_aug}
\end{equation}
\begin{equation}
   \mathcal{L}_{aug} = \mathcal{L}_{sim} + \mathcal{L}_{fst} + \mathcal{L}_{KL} ,
   \label{eq:loss_aug}
\end{equation}
where $\boldsymbol{\theta}^{aug}$ denotes learnable parameters for temporal data augmentation module and $\mathcal{L}_{KL}$ denotes the commonly used regularization term for the adopted VAE~\cite{verma2019manifold}. 
%
% Intuitively, the augmentation is optimized to capture the temporal dependencies learned by the STGNN backbone, and thus could generate temporal data on the same manifold of the original data. 

In \textit{Phase 2}, we aim to jointly optimize the learnable parameters in the STGNN backbone and the spatial topology learning module, where the parameters of temporal data augmentation are frozen. Formally, we have
\begin{equation}
    \boldsymbol{\theta}^{gf}_{\text{epoch + 1}} = \boldsymbol{\theta}^{gf}_{\text{epoch}} - \eta\cdot \nabla_{\boldsymbol{\theta}^{gf}}\mathcal{L}_{gf},
    \label{eq:update_gf}
\end{equation}
\begin{equation}
\begin{aligned}
    \mathcal{L}_{gf} = \mathcal{L}_{fst} + \mathcal{L}_{ori}
    \label{eq:loss_gf}
\end{aligned}
\end{equation}

\noindent where $\boldsymbol{\theta}^{gf}$ denotes the learnable parameters associated with the STGNN backbone and spatial topology learning module. $\mathcal{L}_{fst}$ and $\mathcal{L}_{ori}$ are loss functions of the forecasting results for the generated and original time series, respectively.
In this way, the parameters in the spatial topology learning module and the STGNN backbone are jointly optimized.

% \begin{equation}
% \min _{\theta_{gf}} \mathcal{L}_{gf}(\theta_{gf}, \theta_{aug}^*) \text { such that } \theta_{aug}^*\in \arg \min _{\theta_{aug}}\mathcal{L}_{aug}(\theta_{gf}, \theta_{aug})
% \end{equation}

% \subsection{Inductive Inference}
% temporal data from all nodes is available, which eliminates the need for generating additional temporal data. As a result, we only sample a graph with spatial topology learner and input temporal data. Note that ST-FiT applies to any number of inference nodes due to its inductive capability.
% \noindent \textbf{Inference.}
Finally, during inference, we propose to sample a graph topology characterized by $\tilde{\boldsymbol{A}}$ based on the learned $\boldsymbol{P}$. As such, we are able to directly perform forecasting. We present the complete algorithmic routine in Appendix.

\section{Experimental evaluations}
\label{exp_sec}

In this section, we aim to answer the following research questions. \textbf{RQ1.} How well can ST-FiT generalize to nodes with no available temporal data for training, compared to other existing alternatives? \textbf{RQ2.} How does the performance tendency of ST-FiT look like compared with other baselines when these models are trained on varying ratios of nodes with available temporal data? \textbf{RQ3.} How does each module of ST-FiT contribute to the overall performance? \textbf{RQ4.} How does the choice of hyper-parameters influence the performance of ST-FiT?
In the following sections, we first present the experimental settings, followed by the answers to the proposed research questions.

\subsection{Experimental settings}
Below we provide a brief introduction to the experiment settings, the details will be explained in Appendix. 

\noindent \textbf{Datasets.}
Following previous works~\cite{li2023dmgf}, we conduct experiments on three most commonly used real-world datasets PEMS03, PEMS04, and PEMS08, which are all public transport network
datasets released by Caltrans Performance Measurement
System (PeMS)~\cite{pems2021caltrans}. 

% The three datasets have been uniformly configured to a time granularity of 5 minutes, which is in alignment with prior studies~\cite{li2022automated,li2017diffusion,shao2022pre}. The adjacency matrix of each dataset is derived from the spatial topology of the road network. The Z-score standardization technique~\cite{colan2013and} is utilized to normalize the volume of the traffic flow. The statistics of all three datasets are shown in Table~\ref{tab:datasets}.

% \begin{table}[h]
% \centering
% \caption{Statistics of the adopted real-world Datasets.}
% \label{tab:datasets}
% \setlength{\tabcolsep}{12.65pt}
% \renewcommand{\arraystretch}{1.05}
% \begin{tabular}{cccc}
% \toprule
% Datasets   & \#Nodes & \#Edges & \#TimeSteps    \\
% \midrule 
% PEMS03     & 358  &  547 & 26208 \\
% PEMS04     & 307  &  340 & 16992 \\
% PEMS08     & 170  &  295 & 17856 \\
% \bottomrule
% \end{tabular}
% \end{table}

\noindent\textbf{Baselines.}
Since the studied setting is novel, which requires the model to be inductive, we compare our framework to state-of-the-art baselines applicable to such experimental setting.~\textbf{Linear Sum:} \textit{(1)~Historical Average~(HA)}~\cite{dai2020hybrid}.~\textbf{Temporal-based}: \textit{(2) FC-LSTM}~\cite{sutskever2014sequence}.~\textbf{Spatial-Temporal}: \textit{(3) STGCN}~\cite{yu2017spatio}. \textit{(4) STGODE}~\cite{fang2021spatial}. \textbf{Fine-tuning:} \textit{(5) TransGTR}~\cite{jin2023transferable}.

\begin{table}[t]
% \vspace{4mm}
\setlength{\tabcolsep}{3.65pt}
\renewcommand{\arraystretch}{1.15}
\centering
\caption{Average performance of forecasting. The best results and the second best results are in bold and underlined, respectively. All experiments have been repeated with 3 different random seeds. ST-FiT outperforms baselines without fine-tuning on all datasets, and achieves competitive performance with fine-tuning baseline TransGTR.}
\label{tab:main}
\resizebox{0.9\columnwidth}{!}{
\begin{tabular}{
  l % Dataset column
  l % Methods column
  ccc % Horizon 12 columns
}
\toprule
{Datasets} & {Methods} & {MAE} & {RMSE} & {MAPE (\%)} \\
\midrule
\multirow{6}{*}{PEMS03} & HA & 32.47~\footnotesize{$(\pm 0.00)$}	& 49.80~\footnotesize{$(\pm 0.00)$}	& 30.59~\footnotesize{$(\pm 0.00)$}\\
 & FC-LSTM & 20.56~\footnotesize{$(\pm 0.06)$}	& 33.96~\footnotesize{$(\pm 0.36)$}	& 20.41~\footnotesize{$(\pm 0.40)$}\\
 & STGODE & 31.05~\footnotesize{$(\pm 1.75)$}	& 53.23~\footnotesize{$(\pm 7.95)$}	& 30.20~\footnotesize{$(\pm 1.26)$}\\
         % & FC-LSTM~(\textbf{Ours})       & 22.14 & 35.76& 11.96\% & 24.68 & 39.99 & 13.18\% & 28.53 & 45.88 & 15.53\% \\
         % Add other rows of PEMS-BAY dataset
 & STGCN & 23.22~\footnotesize{$(\pm 0.24)$}	& 37.70~\footnotesize{$(\pm 1.15)$}	& 22.91~\footnotesize{$(\pm 0.80)$}\\
 \cmidrule{2-5}
 & TransGTR & \textbf{17.50}~\footnotesize{$(\pm 0.78)$}	& \textbf{28.35}~\footnotesize{$(\pm 1.16)$}	& \textbf{18.11}~\footnotesize{$(\pm 0.75)$} \\
 \cmidrule{2-5}
 & \textbf{ST-FiT} & \underline{18.40}~\footnotesize{$(\pm 0.23)$}	& \underline{29.31}~\footnotesize{$(\pm 0.32)$}	& \underline{18.94}~\footnotesize{$(\pm 1.53)$}
 \\
 % \addlinespace
 \cmidrule{1-5}
 \multirow{6}{*}{PEMS04} & HA  & 41.98~\footnotesize{$(\pm 0.00)$}	& 61.50~\footnotesize{$(\pm 0.00)$}	& 29.92~\footnotesize{$(\pm 0.00)$}\\
 & FC-LSTM& \underline{28.17}~\footnotesize{$(\pm 0.32)$}	& \underline{44.38}~\footnotesize{$(\pm 0.46)$}	& \underline{19.21}~\footnotesize{$(\pm 0.38)$}\\
 & STGODE & 34.35~\footnotesize{$(\pm 1.62)$}	& 51.54~\footnotesize{$(\pm 1.98)$}	& 25.59~\footnotesize{$(\pm 3.03)$}\\
         % & FC-LSTM~(\textbf{Ours})       & 22.14 & 35.76& 11.96\% & 24.68 & 39.99 & 13.18\% & 28.53 & 45.88 & 15.53\% \\
         % Add other rows of PEMS-BAY dataset
 & STGCN& 32.60~\footnotesize{$(\pm 0.20)$}	& 48.89~\footnotesize{$(\pm 0.74)$}	& 23.40~\footnotesize{$(\pm 0.58)$}\\
 \cmidrule{2-5}
 & TransGTR & 32.76~\footnotesize{$(\pm 2.30)$}	& 48.94~\footnotesize{$(\pm 5.86)$}	& 26.87~\footnotesize{$(\pm 2.54)$}\\
 \cmidrule{2-5}
 & \textbf{ST-FiT}& \textbf{25.11}~\footnotesize{$(\pm 0.42)$}	& \textbf{39.30}~\footnotesize{$(\pm 0.62)$}	& \textbf{17.23}~\footnotesize{$(\pm 0.43)$}\\

% \addlinespace % Space before the next dataset
\cmidrule{1-5}
 \multirow{6}{*}{PEMS08} & HA     & 34.56~\footnotesize{$(\pm 0.00)$}	& 50.41~\footnotesize{$(\pm 0.00)$}	& 21.60~\footnotesize{$(\pm 0.00)$}\\
 & FC-LSTM & 30.52~\footnotesize{$(\pm 0.78)$}	& 49.58~\footnotesize{$(\pm 1.74)$}	& 16.33~\footnotesize{$(\pm 0.30)$}\\
        % & FC-LSTM~(\textbf{Ours})   & 21.61 & 34.21 & 14.81\% & 23.76 & 37.54  & 16.16\% & 27.99 & 44.01 & 19.11\% \\
         % Add other rows of METR-LA dataset
& STGODE & 30.00~\footnotesize{$(\pm 2.10)$}	& 48.24~\footnotesize{$(\pm 5.23)$}	& 18.62~\footnotesize{$(\pm 0.55)$}\\
& STGCN & 41.67~\footnotesize{$(\pm 1.25)$}	& 63.48~\footnotesize{$(\pm 1.37)$}	& 33.48~\footnotesize{$(\pm 3.80)$}\\
\cmidrule{2-5}
& TransGTR & \textbf{18.00}~\footnotesize{$(\pm 0.61)$}	& \textbf{28.32}~\footnotesize{$(\pm 0.79)$}	& \textbf{11.30}~\footnotesize{$(\pm 0.99)$}\\
\cmidrule{2-5}
& \textbf{ST-FiT}& \underline{25.09}~\footnotesize{$(\pm 0.18)$}	& \underline{39.52}~\footnotesize{$(\pm 0.46)$}	& \underline{14.48}~\footnotesize{$(\pm 0.47)$}\\
 % Space before the next dataset
 
\bottomrule
\end{tabular}
}
\end{table}

\noindent\textbf{Task Settings.} 
For a fair comparison, we follow the dataset division along temporal dimensions in previous works~\cite{jiang2023enhancing,liu2023spatio}, where datasets are split as 70\% training, 20\% validation and 10\% inference in chronological order. 
% where we split the first 60\% temporal data for training, the subsequent 20\% temporal data for validation, and the remaining 20\% temporal data for inference. 
For the task setting of inductive forecasting with limited training data, we randomly choose the temporal data from 10\% nodes for training, and adopt the same split for validation. 
Following previous works, we generate training samples through a sliding window of 24 time steps, with the first 12 as model input, and the remaining 12 as ground truth for forecasting outcomes. Accordingly, we compare the average performance on the MAE, RMSE, and MAPE metrics.

\begin{figure*}
    \centering
    \begin{subfigure}[t]{0.33\textwidth} % subfigure环境，[b]表示对齐方式，0.48\textwidth设置宽度
        \centering
        \includegraphics[width=\textwidth]{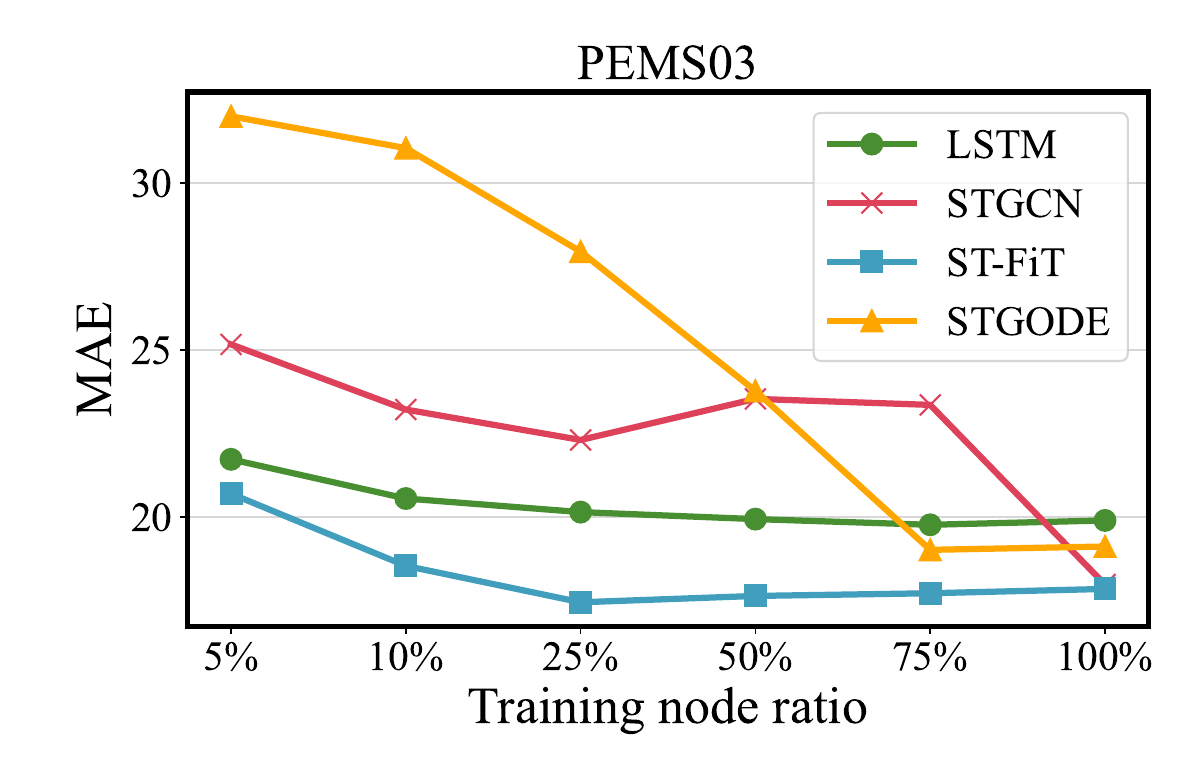} % 插入图形
        \caption{MAE for PEMS03} % 子图的标题
        \label{fig:ratio_a} % 子图的标签，用于交叉引用
    \end{subfigure}
    \hfill % 横向填充，确保子图之间有适当间隔
    \begin{subfigure}[t]{0.33\textwidth}
        \centering
        \includegraphics[width=\textwidth]{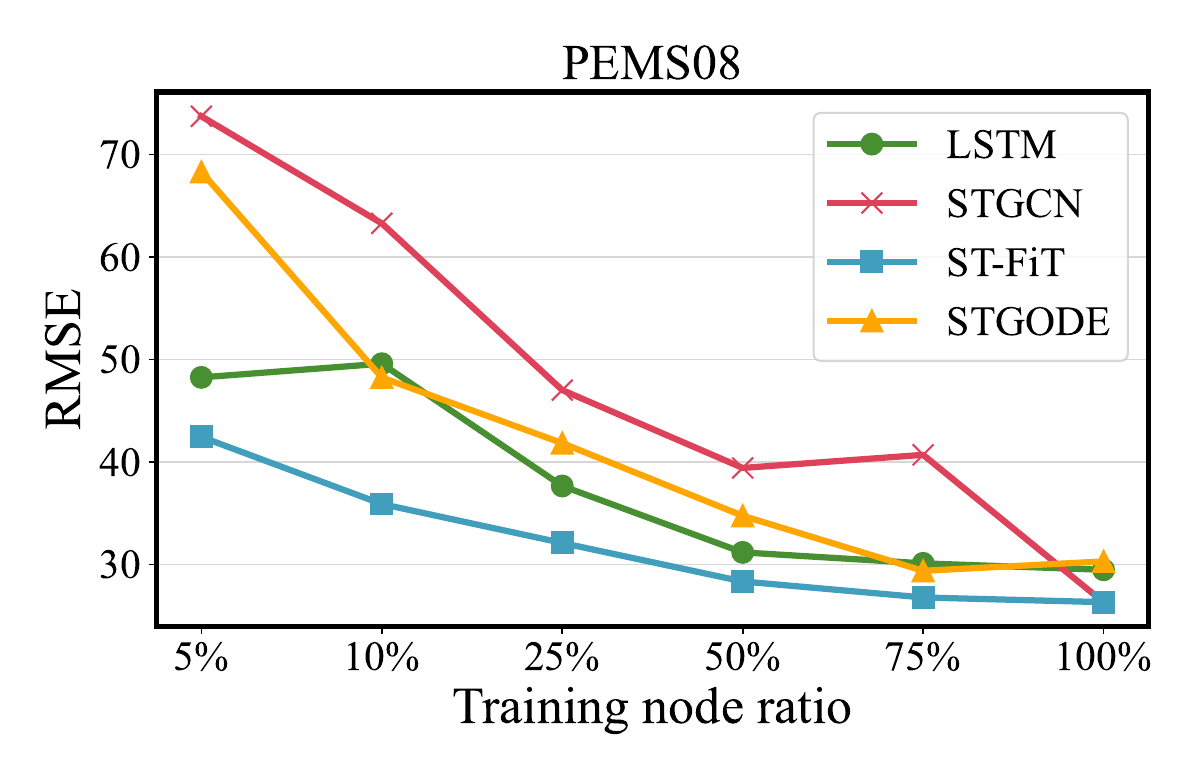}
        \caption{RMSE for PEMS08}
        \label{fig:ratio_b}
    \end{subfigure}
    \hfill % 横向填充，确保子图之间有适当间隔
    \begin{subfigure}[t]{0.33\textwidth}
        \centering
        \includegraphics[width=\textwidth]{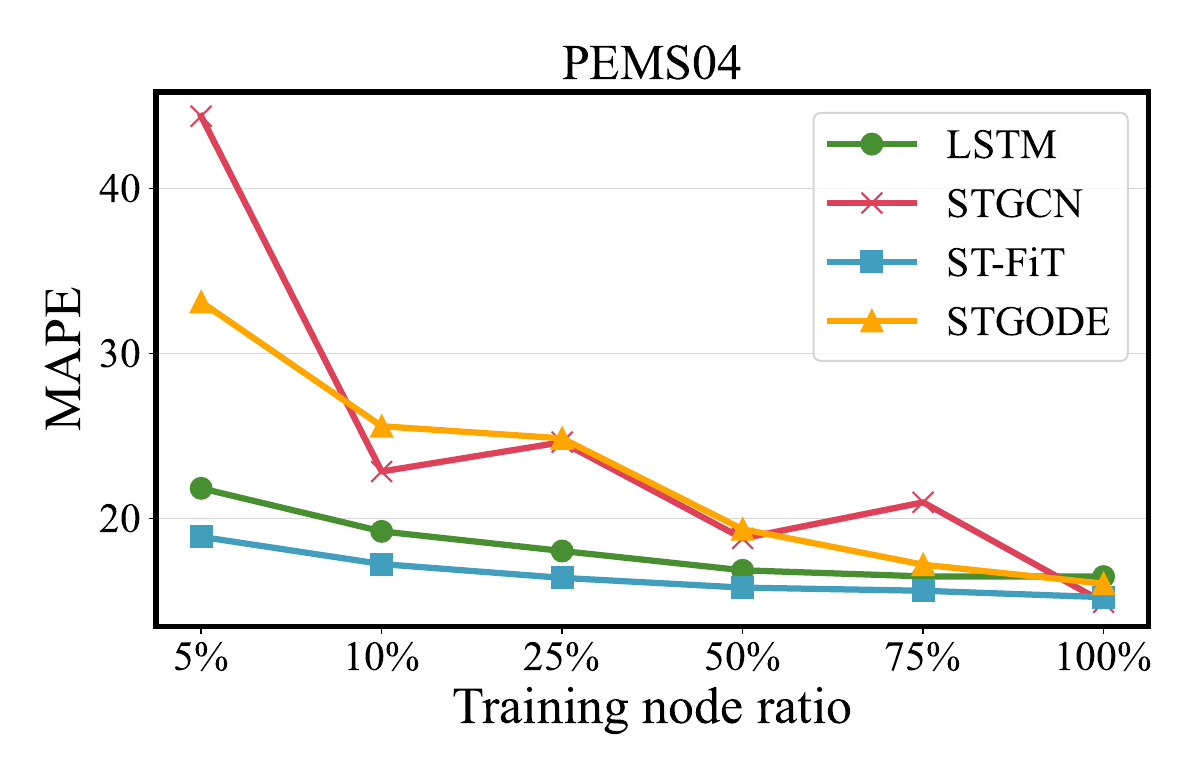}
        \caption{MAPE for PEMS04}
        \label{fig:ratio_c}
    \end{subfigure}
    % \vspace{-2mm}
    \caption{The performance of ST-FiT compared to baselines with different training node ratios. As training node ratio decreases, the performances of all models drops, while ST-FiT outperforms other baselines over all ratios consistently.}
    \label{fig:4}
\end{figure*}

\subsection{Generalization Performance}

% in its ability to generalize to different nodes and compare its performance with common baselines with or without fine-tuning. The results on three datasets are shown in Table~\ref{tab:main}. We make the following observations: 

To answer \textbf{RQ1}, we first evaluate the forecasting performance of ST-FiT on those nodes without available temporal data during training. We make following observations from the the empirical results in Table~\ref{tab:main}.
(1) ST-FiT outperforms all baselines that do not require fine-tuning. Notably, ST-FiT exceeds the associated backbone model STGCN by up to 40.0\% in MAE, 38.6\% in RMSE, and 55.6\% in MAPE. This verifies the effectiveness of ST-FiT in generalizing to the nodes with different temporal dependencies. Meanwhile, ST-FiT also significantly outperforms FC-LSTM and STGODE, which further demonstrates its superiority over different types of state-of-the-art alternatives. (2) ST-FiT achieves comparable performance with fine-tuned model TransGTR. Specifically, ST-FiT outperforms TransGTR on PEMS04 by 27.1\% in MAE, 19.0\% in RMSE, and  34.8\% in MAPE. In addition, ST-FiT has comparable performances with TransGTR on PEMS03. We note that compared with ST-FiT, TransGTR adopts an additional fine-tuning process with much more abundant temporal data. As such, TransGTR can extract temporal data from all nodes in the fine-tuning process, which helps it capture more spatial-temporal dependencies. As a result, we argue that achieving competitive performance to TransGTR without relying on any fine-tuning process should be regarded as satisfactory performance, and this further corroborates the effectiveness of ST-FiT in capturing diverse spatial-temporal dependencies.

\subsection{Performance w.r.t. Training Node Ratios}
To answer \textbf{RQ2}, we train the proposed ST-FiT with different ratios of nodes whose temporal data is available during training. We compare its performance tendencies with those from baselines without fine-tuning. We select a wide range of ratios including 5\%, 10\%, 25\%, 50\%, 75\%, and 100\% to test whether ST-FiT can outperform other aternatives consistently w.r.t. different ratios. 
% Additionally, we also incorporate the performance with a ratio of 100\% nodes for comparison.
The results are shown in Figure~\ref{fig:4}. We make the following observations: (1) ST-FiT consistently outperforms baselines on all ratios, which demonstrates its effectiveness under different degrees of training data limitation. (2) ST-FiT performs better with fewer training nodes. For PEMS08, we could observe 28.1\% improvement in RMSE compared to STGCN with 50\% ratio, while it increases to 42.4\% with 5\% ratio.
% For PEMS08, when the train node ratio is 50\%, we could observe at most 28.1\% improvements in RMSE compared to STGCN, while train node ratio drops to 5\%, we could observe at most 42.4\% improvements in RMSE. 
This observation demonstrates ST-FiT's effectiveness especially when training with severely limited data. (3) When all nodes have temporal data for training, ST-FiT still remains comparable with the best baselines, which indicates our temporal data augmentation and spatial topology learning do not impair the STGNN backbone's original performance and are widely applicable.
% and can be applied to a broader range of scenarios.

\begin{table}[!t]
% \vspace{4mm}
\setlength{\tabcolsep}{3.65pt}
\renewcommand{\arraystretch}{1.15}
\centering
\caption{Performance comparison of forecasting for ablation study. The best results are in bold, and the second best results are underlined. It is observed that removing any module of ST-FiT will jeopardize the overall performance.}
\label{tab:ablation}
\resizebox{0.9\columnwidth}{!}{
\begin{tabular}{
  l % Methods column
ccc % Horizon 3 columns
}
\toprule
 {Variants}  & {MAE} & {RMSE} & {MAPE (\%)} \\
\midrule
\textbf{ST-FiT} & \textbf{25.11}~\footnotesize{$(\pm 0.42)$}	& \textbf{39.30}~\footnotesize{$(\pm 0.62)$}	& \textbf{17.23}~\footnotesize{$(\pm 0.43)$}\\
 \midrule
 {w/o aug}	& 27.31~\footnotesize{$(\pm 0.18)$}	& 43.04~\footnotesize{$(\pm 0.29)$}	& 18.64~\footnotesize{$(\pm 0.10)$}\\
 {w/o sim}	& \underline{25.55}~\footnotesize{$(\pm 0.14)$}	& \underline{40.18}~\footnotesize{$(\pm 0.44)$}	& \underline{17.41}~\footnotesize{$(\pm 0.16)$}\\

 {w/o fst}	& 26.61~\footnotesize{$(\pm 1.06)$}	& 41.66~\footnotesize{$(\pm 1.61)$}	& 19.00~\footnotesize{$(\pm 0.56)$}\\
  \midrule
 {w/o gl}& 27.58~\footnotesize{$(\pm 0.30)$}	& 43.50~\footnotesize{$(\pm 0.55)$}	& 19.00~\footnotesize{$(\pm 0.28)$}   \\
 {w/o gs}	& 26.68~\footnotesize{$(\pm 1.12)$}	& 42.55~\footnotesize{$(\pm 2.20)$}	& 17.76~\footnotesize{$(\pm 0.35)$}\\
 {identity}&  28.00~\footnotesize{$(\pm 0.18)$}	& 44.48~\footnotesize{$(\pm 0.36)$}	& 19.38~\footnotesize{$(\pm 0.60)$}\\

% \addlinespace % Space before the next dataset

% \addlinespace % Space before the next dataset
 
\bottomrule
\end{tabular}
}
\end{table}

% \begin{figure*}
%     \centering
%     \includegraphics[width=\textwidth]{src/ratio.pdf}
%     \caption{The performance of ST-FiT compared to baselines with different training node ratios. As training node ratio decreases, the performances of all models drops, while ST-FiT outperforms other baselines over all ratios consistently.}
%     \label{fig:4}
% \end{figure*}

% \begin{figure*}
%     \centering
%     \begin{subfigure}[t]{0.33\textwidth} % subfigure环境，[b]表示对齐方式，0.48\textwidth设置宽度
%         \centering
%         \includegraphics[width=\textwidth]{src/ratio_a.pdf} % 插入图形
%         \caption{MAE for PEMS03} % 子图的标题
%         \label{fig:ratio_a} % 子图的标签，用于交叉引用
%     \end{subfigure}
%     \hfill % 横向填充，确保子图之间有适当间隔
%     \begin{subfigure}[t]{0.33\textwidth}
%         \centering
%         \includegraphics[width=\textwidth]{src/ratio_b.pdf}
%         \caption{RMSE for PEMS08}
%         \label{fig:ratio_b}
%     \end{subfigure}
%     \hfill % 横向填充，确保子图之间有适当间隔
%     \begin{subfigure}[t]{0.33\textwidth}
%         \centering
%         \includegraphics[width=\textwidth]{src/ratio_c.pdf}
%         \caption{MAPE for PEMS04}
%         \label{fig:ratio_c}
%     \end{subfigure}
%     % \vspace{-2mm}
%     \caption{The performance of ST-FiT compared to baselines with different training node ratios. As training node ratio decreases, the performances of all models drops, while ST-FiT outperforms other baselines over all ratios consistently.}
%     \label{fig:4}
% \end{figure*}

\subsection{Ablation Study}
To answer \textbf{RQ3}, we investigate how our proposed modules contribute to the forecasting performance separately. We use \textit{w/o aug} to denote cases without temporal data augmentation module. \textit{W/o gl} denotes removing the spatial topology learning module. Furthermore, we design two experiments for temporal data augmentation, where \textit{w/o sim} denotes removing $\mathcal{L}_{sim}$, and \textit{w/o fst} denotes removing $\mathcal{L}_{fst}$. Additionally, we design two variants of our proposed spatial topology learning module. specifically, \textit{w/o gs} denotes adopting a fully connected graph. \textit{Identity} denotes replacing $\boldsymbol{A}$ with an identity matrix. The results are shown in Table~\ref{tab:ablation}. We make the observations as follows: (1) Both modules contribute to the overall performance, which verifies their effectiveness for improving generalization. (2) Removing $\mathcal{L}_{sim}$ or $\mathcal{L}_{fst}$ degrades the performance, which indicates their ability in generating temporal data with consistent temporal dependencies. (3) Spatial topology learning performs better than variant, which indicates the learned spatial topology significantly enhances the diversity of spatial dependencies.

\subsection{Parameter Sensitivity}
\begin{figure}
    \centering
    \begin{subfigure}[t]{0.49\columnwidth} % subfigure环境，[b]表示对齐方式，0.48\textwidth设置宽度
        \centering
        \includegraphics[width=\columnwidth]{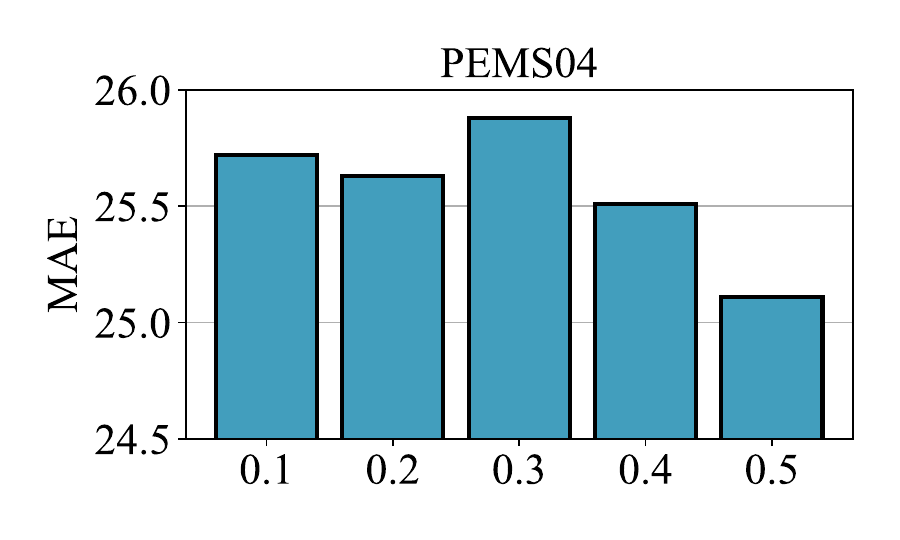} % 插入图形
        \caption{MAE for mix-up ratio $\lambda$} % 子图的标题
        \label{fig:f5_a} % 子图的标签，用于交叉引用
    \end{subfigure}
    \hfill % 横向填充，确保子图之间有适当间隔
    \begin{subfigure}[t]{0.49\columnwidth}
        \centering
        \includegraphics[width=\columnwidth]{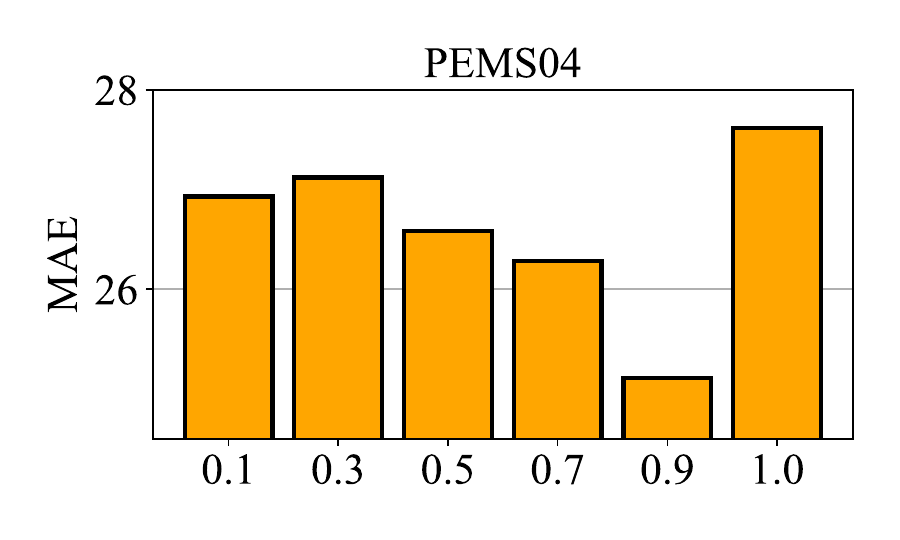}
        \caption{MAE for sparse threshold $\epsilon$}
        \label{fig:f5_b}
    \end{subfigure}
    \begin{subfigure}[t]{0.49\columnwidth} % subfigure环境，[b]表示对齐方式，0.48\textwidth设置宽度
        \centering
        \includegraphics[width=\columnwidth]{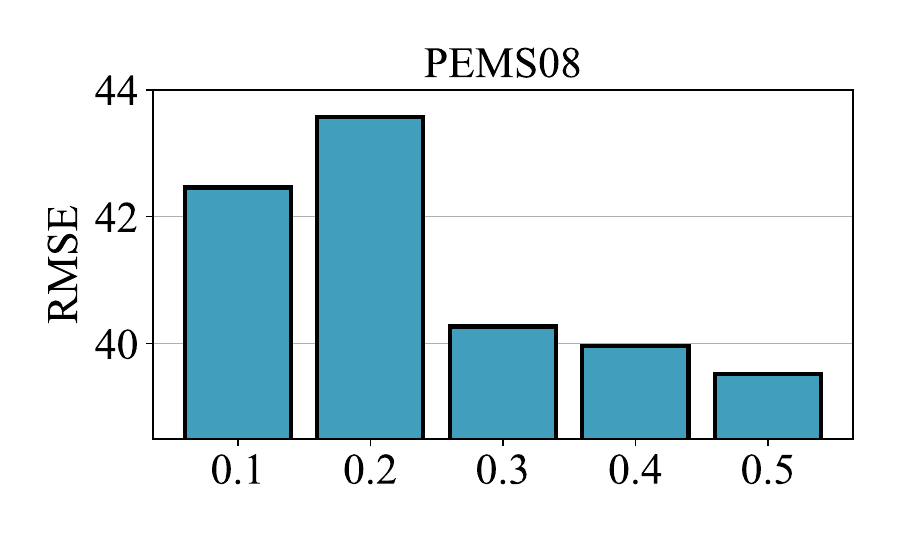} % 插入图形
        \caption{RMSE for mix-up ratio $\lambda$} % 子图的标题
        \label{fig:f5_c} % 子图的标签，用于交叉引用
    \end{subfigure}
    \hfill % 横向填充，确保子图之间有适当间隔
    \begin{subfigure}[t]{0.49\columnwidth}
        \centering
        \includegraphics[width=\columnwidth]{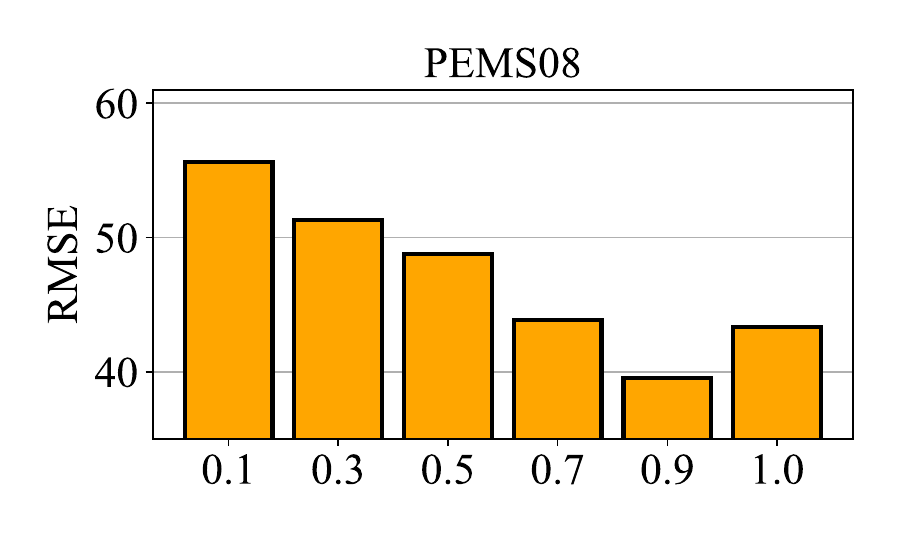}
        \caption{RMSE for sparse threshold $\epsilon$ }
        \label{fig:f5_d}
    \end{subfigure}
    \vspace{-2mm}
    \caption{Performance of ST-FiT with different mix-up ratios $\lambda$ and sparse thresholds $\o$. The mix-up ratio behaves slight influence, while the positive correlation between mix-up ratio $\lambda$ and the performance still exhibits. Sparsity has positive correlation with performance, while extreme sparsity brings negative influence due to loss of key connections.}
    \label{fig:5}
\end{figure}

To answer \textbf{RQ4}, we analyze the impact of hyper-parameter values, including the threshold in controlling sparsity of spatial topology learning $\epsilon$ and the parameter $\lambda$ for temporal data mix-up. We choose values of $\epsilon$ from 0 to 1, where higher value denotes a sparser learned graph topology. For the value of $\lambda$, we choose it from the range between 0 and 0.5. The results of three datasets are shown in Figure \ref{fig:5}. For the impact of the sparsity of the graph, we observe that: (1) Sparser structures generally perform better, which indicates the necessity of sparsity in spatial-temporal forecasting on nodes without available training temporal data. (2) Overly sparse structures harm performance, since it can omit certain key connections between nodes. From Figure \ref{fig:f5_a} and \ref{fig:f5_b}, we are able to observe that the higher value of $\lambda$ improve the performance, which can be attributed to a higher diversity of the generated temporal data. With above experiments, we recommend using $\lambda$ as 0.5, $\epsilon$ as 0.9 for optimal performance.

% \begin{figure}
%     \centering
%     \includegraphics[width=\columnwidth]{src/lambda_sparse.pdf}
%     \caption{Performance of ST-FiT with different mix-up ratios $\lambda$ and sparse thresholds $\o$. The mix-up ratio behaves slight influence, while the positive correlation between mix-up ratio $\lambda$ and the performance still exhibits. Sparsity has positive correlation with performance, while extreme sparsity might bring negative influence due to loss of core connections.}
%     \label{fig:5}
% \end{figure}

\section{Related Works}
\noindent\textbf{Spatial-Temporal Forecasting.} Spatial-temporal forecasting is crucial but challenging. STGNNs have shown promise in this area but typically require temporal data for all nodes during training, making them unsuitable for inductive forecasting~\cite{deng2021st}. Some other STGNNs can perform inductive forecasting~\cite{wu2019graph,jiang2023spatio} but struggle with limited training data, where they only extract limited spatial-temporal dependencies and have difficulty adapting to temporal data with new spatial-temporal dependencies.
Recently, several works have resorted to domain adaptation for such generalization challenges. They managed to extract temporal dependencies from nodes without temporal data during training, which offers a possible solution to inductive forecasting~\cite{cheng2023weakly,ouyang2023citytrans} on limited training data.  However, these works focus on graph-level generalization and overlook granular temporal dependency differences between nodes in the same spatial-temporal graph. In addition, they require costly fine-tuning, which limits their efficiency for some inductive forecasting scenarios. In contrast, ST-FiT enriches the training data with diverse temporal dependencies through temporal data augmentation and captures different spatial dependencies between existing and new temporal data, which helps the generalization to different nodes.

\noindent\textbf{Temporal Data Augmentation.}
Due to the common scarcity of temporal data in real-world scenarios, existing works propose to generate temporal data with data augmentation techniques~\cite{fu2020data}. The most challenging part is to generate data not only with diverse temporal dependencies but also lying close to the manifold of existing temporal data. Most traditional algorithms such as slicing, jittering, or scaling apply simple transformation and perturbation~\cite{um2017data,iwana2021time}, where they either fail to generate temporal data with diverse and consistent temporal dependencies. Recent prevailing deep learning models such as Generative Adversarial Networks (GANs)~\cite{goodfellow2020generative} and Variational Autoencoders (VAEs)~\cite{goubeaud2021using} are promising solutions to generating more consistent temporal data that lie close to the manifold of existing temporal data.
% since they augment on the learned manifold which could better capture the temporal dependencies of data. 
Nevertheless, the generated temporal data lack diversity, which limits the contribution to generalization ability.
% Recently, et al. verifies that augmenting the latent space with the mix-up could enrich the latent space regions where data might be missing, which is especially useful for limited training data. Inspired by this work, we 
% Another line of temporal data augmentation is temporal mix-up~\cite{demirel2024finding,wickstrom2022mixing}, where generated temporal data is diverse but might generate data that greatly deviates from the manifold of the original temporal data, due to the simple convex combination of two time series with low correlations. 
%
% Incorporating both advantages of generative models and mix-up,
In this work, we present ST-FiT, which applies mix-up on the manifold and learns to capture the temporal dependencies. This helps to not only enrich the original latent space region due to mix-up on the manifold~\cite{huh2024isometric}, but also ensure that the generated temporal data lies close to the manifold where the available training temporal data lies.

%现在还没有形成一个比较deep的习惯，而且那不是一个思考的状态。他们的缺点：
\noindent\textbf{Graph Topology Learning.}
Due to incompleteness and scarcity of existing graph topology, substantial work have devoted to learning better graph topology over diverse types of network data such as brain networks~\cite{cui2022braingb}, social networks~\cite{zhang2022robust}, and financial transaction networks~\cite{wang2021risk}. One of the most important problems is to learn a sparse and discrete graph topology that not only represents the real-world scenarios but also contains few spurious connections~\cite{jin2020graph}. \cite{franceschi2019learning} proposed to leverage Gumbel-Softmax reparameterization tricks in learning discrete structures with bilevel optimization.
% where the sparsity is achieved by regularizing the learned structure on a pre-defined graph which is constructed from the $k$-nearest neighbor ($k$NN). 
\cite{shang2021discrete} further applied Gumbel-Softmax reparameterization to spatial-temporal forecasting with uni-level optimization.
However, they both require additional resources such as a pre-defined KNN to achieve topology sparsity.
% and indicated a uni-level optimization of both graph and forecasting model could achieve satisfying performance. 
In contrast, ST-FiT employs a simple transformation to attain any desired level of sparsity without increasing the computational burden.
% , which equips the forecasting model with generalization capability across different spatial dependencies.

\section{Conclusion}
In this paper, we study an under-explored research problem of inductive forecasting with limited training data, which requires models to generalize the learned spatial-temporal dependencies from the nodes with available training temporal data to those nodes without. To handle this problem, we propose ST-FiT that can achieve superior performance without additional fine-tuning. Overall, two key learning modules contribute to the superiority of ST-FiT.
Specifically, the temporal data augmentation allows us to generate diverse temporal data which lies close to the manifold of the training temporal data. Hence the model is enabled to generalize to nodes with different temporal dependencies. 
Meanwhile, the spatial topology learning refines the spatial dependencies, which improves adaptation to the backbone STGNNs to achieve better forecasting performance. 
We also propose an iterative training strategy for the optimization of these modules. Extensive experiments on three commonly used real-world datasets verify the effectiveness of ST-FiT.

% \appendix
% \section{Reference Examples}
% \label{sec:reference_examples}
\section{Acknowledgments}
This work is supported in part by the National Science Foundation under grants (IIS-2006844, IIS-2144209, IIS-2223769, IIS-2331315, CNS-2154962, BCS-2228534, and CMMI-2411248), Office of Naval Research under grant (N000142412636), the Commonwealth Cyber Initiative Awards under grants (VV-1Q24-011, VV-1Q25-004), and the research gift funding from Netflix and Snap.

\bibliography{aaai24}

\appendix

% \newpage
\section{Reproducibility}
In this section, we introduce the details of the experiments in this paper for the purpose of reproducibility.
% At the same time, we have uploaded all necessary code to the \textbf{Anonymous GitHub}\footnote{https://anonymous.4open.science/} to reproduce the results presented in this paper: \textcolor{blue}{\url{https://anonymous.4open.science/r/InductiveST-521F/}}.
All major experiments are encapsulated as shell scripts, which can be easily executed. We introduce details in subsections below.
\begin{figure*}[!h]
    \centering
    \includegraphics[width=\textwidth]{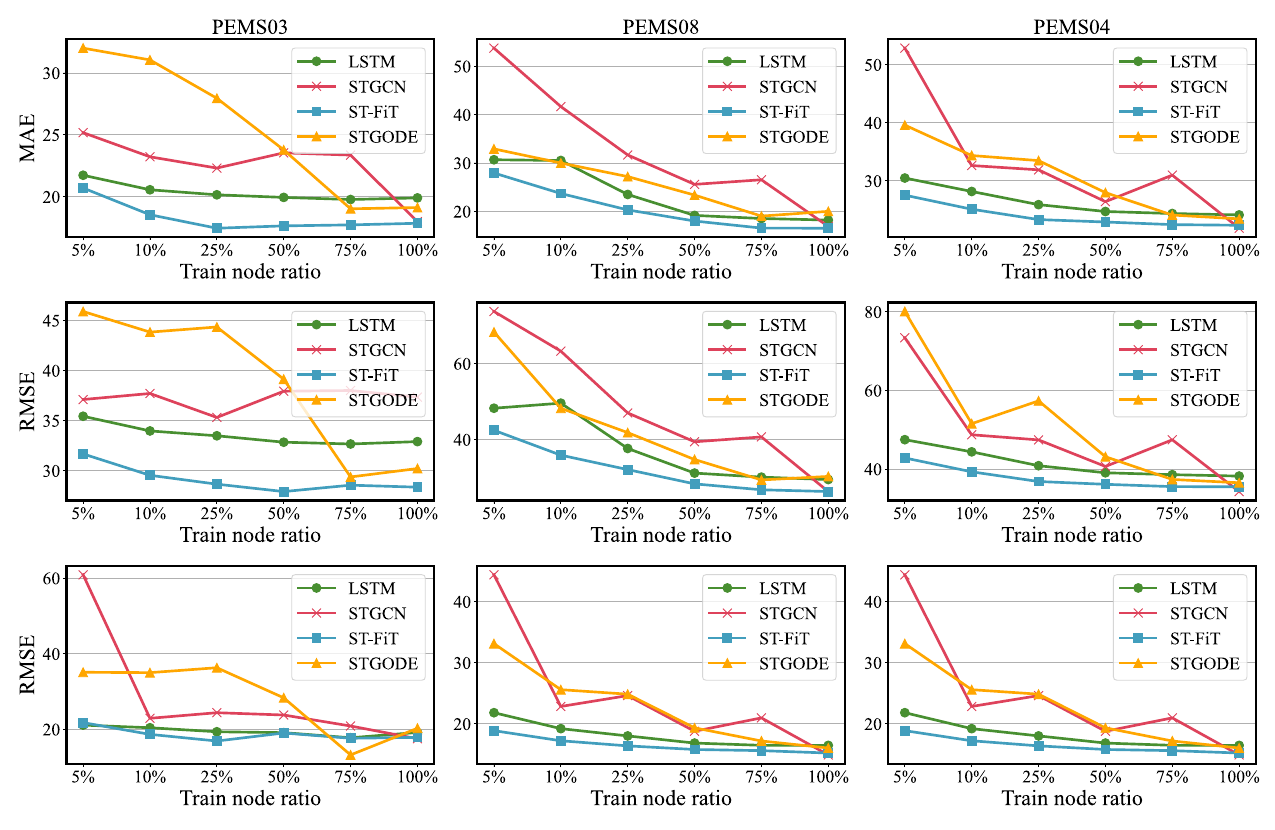}
    \caption{The performance of ST-FiT compared to baselines with different training node ratios. As training node ratio decreases, performances of all models drops, while ST-FiT outperforms other baselines over all ratios consistently.}
    \label{fig:2}
\end{figure*}

\subsection{Dataset Details}
we conduct experiments on three most commonly used real-world datasets PEMS03, PEMS04, and PEMS08, which are all public transport network
datasets released by Caltrans Performance Measurement
System (PeMS)~\cite{pems2021caltrans}. The three datasets have been uniformly configured to a time granularity of 5 minutes, which is in alignment with prior studies~\cite{li2022automated,shao2022pre}. The adjacency matrix of each dataset is derived from the spatial topology of the road network. The Z-score standardization technique~\cite{colan2013and} is utilized to normalize the volume of the traffic flow. The statistics of all three datasets are shown in Table~\ref{tab:datasets}.

\begin{table}[h]
\centering
\caption{Statistics of the adopted real-world Datasets.}
\label{tab:datasets}
\setlength{\tabcolsep}{12.65pt}
\resizebox{0.95\columnwidth}{!}{
\renewcommand{\arraystretch}{1.1}
\begin{tabular}{cccc}
\toprule
Datasets   & \#Nodes & \#Edges & \#TimeSteps    \\
\midrule 
PEMS03     & 358  &  547 & 26208 \\
PEMS04     & 307  &  340 & 16992 \\
PEMS08     & 170  &  295 & 17856 \\
\bottomrule
\end{tabular}
}
\end{table}

\begin{table*}[!htbp]
\vspace{4mm}
\setlength{\tabcolsep}{3.65pt}
\renewcommand{\arraystretch}{1.15}
\centering
\caption{Average performance on 3, 6, 12 horizons of forecasting. The best results and the second best results are in bold and underlined, respectively. All experiments have been repeated with 3 different random seeds. ST-FiT outperforms baselines without fine-tuning on all datasets, and achieves competitive performance with fine-tuning baseline TransGTR.}
\label{tab:main_full}
\resizebox{\textwidth}{!}{
\begin{tabular}{
  l % Dataset column
  l % Methods column
ccc % Horizon 3 columns
  ccc % Horizon 6 columns
  ccc % Horizon 12 columns
}
\toprule
Datasets & {Methods} & \multicolumn{3}{c}{Horizon 3} & \multicolumn{3}{c}{Horizon 6} & \multicolumn{3}{c}{Horizon 12} \\
\cmidrule(lr){3-5}
\cmidrule(lr){6-8}
\cmidrule(lr){9-11}
 & & {MAE} & {RMSE} & {MAPE (\%)} & {MAE} & {RMSE} & {MAPE (\%)} & {MAE} & {RMSE} & {MAPE (\%)} \\
\midrule
\multirow{6}{*}{PEMS03} & HA & 32.48~\footnotesize{$(\pm 0.00)$}	& 49.81~\footnotesize{$(\pm 0.00)$}	& 30.58~\footnotesize{$(\pm 0.00)$}	& 32.48~\footnotesize{$(\pm 0.00)$}	& 49.81~\footnotesize{$(\pm 0.00)$}	& 30.58~\footnotesize{$(\pm 0.00)$}	& 32.47~\footnotesize{$(\pm 0.00)$}	& 49.80~\footnotesize{$(\pm 0.00)$}	& 30.59~\footnotesize{$(\pm 0.00)$}\\
 & FC-LSTM & 15.17~\footnotesize{$(\pm 0.04)$}	& 25.31~\footnotesize{$(\pm 0.25)$}	& 15.92~\footnotesize{$(\pm 0.38)$}	& 17.00~\footnotesize{$(\pm 0.03)$}	& 28.29~\footnotesize{$(\pm 0.30)$}	& 17.19~\footnotesize{$(\pm 0.21)$}	& 20.56~\footnotesize{$(\pm 0.06)$}	& 33.96~\footnotesize{$(\pm 0.36)$}	& 20.41~\footnotesize{$(\pm 0.40)$}\\
 & STGODE & 23.08~\footnotesize{$(\pm 2.28)$}	& 37.34~\footnotesize{$(\pm 1.49)$}	& 24.87~\footnotesize{$(\pm 3.16)$}	& 25.38~\footnotesize{$(\pm 1.23)$}	& 41.66~\footnotesize{$(\pm 0.87)$}	& 26.28~\footnotesize{$(\pm 2.24)$}	& 31.05~\footnotesize{$(\pm 1.75)$}	& 53.23~\footnotesize{$(\pm 7.95)$}	& 30.20~\footnotesize{$(\pm 1.26)$}\\
         % & FC-LSTM~(\textbf{Ours})       & 22.14 & 35.76& 11.96\% & 24.68 & 39.99 & 13.18\% & 28.53 & 45.88 & 15.53\% \\
         % Add other rows of PEMS-BAY dataset
 & STGCN & 17.10~\footnotesize{$(\pm 0.41)$}	& 28.45~\footnotesize{$(\pm 2.47)$}	& 17.18~\footnotesize{$(\pm 0.77)$}	& 19.16~\footnotesize{$(\pm 0.35)$}	& 31.49~\footnotesize{$(\pm 1.92)$}	& 18.97~\footnotesize{$(\pm 0.16)$}	& 23.22~\footnotesize{$(\pm 0.24)$}	& 37.70~\footnotesize{$(\pm 1.15)$}	& 22.91~\footnotesize{$(\pm 0.80)$}\\
 \cmidrule{2-11}
 & TransGTR & \textbf{14.14}~\footnotesize{$(\pm 0.16)$}	& \underline{23.12}~\footnotesize{$(\pm 0.45)$}	& \textbf{15.59}~\footnotesize{$(\pm 0.32)$}	& \textbf{15.34}~\footnotesize{$(\pm 0.30)$}	& \textbf{25.12}~\footnotesize{$(\pm 0.61)$}	& \textbf{16.63}~\footnotesize{$(\pm 0.52)$}	& \textbf{17.50}~\footnotesize{$(\pm 0.78)$}	& \textbf{28.35}~\footnotesize{$(\pm 1.16)$}	& \textbf{18.11}~\footnotesize{$(\pm 0.75)$}\\
  \cmidrule{2-11}
 & \textbf{ST-FiT} & \underline{14.71}~\footnotesize{$(\pm 0.30)$}	& \textbf{22.80}~\footnotesize{$(\pm 0.34)$}	& \underline{16.53}~\footnotesize{$(\pm 1.16)$}	& \underline{16.03}~\footnotesize{$(\pm 0.25)$}	& \underline{25.21}~\footnotesize{$(\pm 0.36)$}	& \underline{17.40}~\footnotesize{$(\pm 1.60)$}	& \underline{18.40}~\footnotesize{$(\pm 0.23)$}	& \underline{29.31}~\footnotesize{$(\pm 0.32)$}	& \underline{18.94}~\footnotesize{$(\pm 1.53)$}\\
 
 % \addlinespace
 \cmidrule{1-11}
 \multirow{6}{*}{PEMS04} & HA  & 41.94~\footnotesize{$(\pm 0.00)$}	& 61.48~\footnotesize{$(\pm 0.00)$}	& 29.89~\footnotesize{$(\pm 0.00)$}	& 41.96~\footnotesize{$(\pm 0.00)$}	& 61.49~\footnotesize{$(\pm 0.00)$}	& 29.90~\footnotesize{$(\pm 0.00)$}	& 41.98~\footnotesize{$(\pm 0.00)$}	& 61.50~\footnotesize{$(\pm 0.00)$}	& 29.92~\footnotesize{$(\pm 0.00)$}\\
 & FC-LSTM& \underline{21.67}~\footnotesize{$(\pm 0.22)$}	& \underline{34.42}~\footnotesize{$(\pm 0.30)$}	& \underline{14.51}~\footnotesize{$(\pm 0.34)$}	& \underline{23.86}~\footnotesize{$(\pm 0.25)$}	& \underline{37.80}~\footnotesize{$(\pm 0.32)$}	& \underline{16.04}~\footnotesize{$(\pm 0.35)$}	& \underline{28.17}~\footnotesize{$(\pm 0.32)$}	& \underline{44.38}~\footnotesize{$(\pm 0.46)$}	& \underline{19.21}~\footnotesize{$(\pm 0.38)$}\\
 & STGODE & 27.96~\footnotesize{$(\pm 1.66)$}	& 41.24~\footnotesize{$(\pm 1.37)$}	& 21.39~\footnotesize{$(\pm 4.29)$}	& 29.80~\footnotesize{$(\pm 1.55)$}	& 44.12~\footnotesize{$(\pm 1.25)$}	& 22.57~\footnotesize{$(\pm 4.00)$}	& 34.35~\footnotesize{$(\pm 1.62)$}	& 51.54~\footnotesize{$(\pm 1.98)$}	& 25.59~\footnotesize{$(\pm 3.03)$}\\
         % & FC-LSTM~(\textbf{Ours})       & 22.14 & 35.76& 11.96\% & 24.68 & 39.99 & 13.18\% & 28.53 & 45.88 & 15.53\% \\
         % Add other rows of PEMS-BAY dataset
 & STGCN& 24.43~\footnotesize{$(\pm 0.31)$}	& 37.05~\footnotesize{$(\pm 0.71)$}	& 17.75~\footnotesize{$(\pm 0.21)$}	& 27.18~\footnotesize{$(\pm 0.22)$}	& 40.95~\footnotesize{$(\pm 0.60)$}	& 19.63~\footnotesize{$(\pm 0.32)$}	& 32.60~\footnotesize{$(\pm 0.20)$}	& 48.89~\footnotesize{$(\pm 0.74)$}	& 23.40~\footnotesize{$(\pm 0.58)$}\\
 \cmidrule{2-11}
 & TransGTR & 27.27~\footnotesize{$(\pm 0.99)$}	& 39.18~\footnotesize{$(\pm 0.78)$}	& 26.71~\footnotesize{$(\pm 6.10)$}	& 29.45~\footnotesize{$(\pm 0.47)$}	& 42.97~\footnotesize{$(\pm 2.45)$}	& 26.80~\footnotesize{$(\pm 4.29)$}	& 32.76~\footnotesize{$(\pm 2.30)$}	& 48.94~\footnotesize{$(\pm 5.86)$}	& 26.87~\footnotesize{$(\pm 2.54)$}\\
  \cmidrule{2-11}
 & \textbf{ST-FiT}& \textbf{20.31}~\footnotesize{$(\pm 0.34)$}	& \textbf{31.97}~\footnotesize{$(\pm 0.54)$}	& \textbf{14.06}~\footnotesize{$(\pm 0.10)$}	& \textbf{21.95}~\footnotesize{$(\pm 0.35)$}	& \textbf{34.48}~\footnotesize{$(\pm 0.55)$}	& \textbf{15.13}~\footnotesize{$(\pm 0.26)$}	& \textbf{25.11}~\footnotesize{$(\pm 0.42)$}	& \textbf{39.30}~\footnotesize{$(\pm 0.62)$}	& \textbf{17.23}~\footnotesize{$(\pm 0.43)$}\\
 
% \addlinespace % Space before the next dataset
\cmidrule{1-11}
 \multirow{6}{*}{PEMS08} & HA     & 34.54~\footnotesize{$(\pm 0.00)$}	& 50.40~\footnotesize{$(\pm 0.00)$}	& 21.56~\footnotesize{$(\pm 0.00)$}	& 34.55~\footnotesize{$(\pm 0.00)$}	& 50.41~\footnotesize{$(\pm 0.00)$}	& 21.58~\footnotesize{$(\pm 0.00)$}	& 34.56~\footnotesize{$(\pm 0.00)$}	& 50.41~\footnotesize{$(\pm 0.00)$}	& 21.60~\footnotesize{$(\pm 0.00)$}\\
 & FC-LSTM & 23.77~\footnotesize{$(\pm 1.02)$}	& 39.03~\footnotesize{$(\pm 2.50)$}	& 12.71~\footnotesize{$(\pm 0.30)$}	& 26.38~\footnotesize{$(\pm 0.90)$}	& 43.29~\footnotesize{$(\pm 2.11)$}	& 14.02~\footnotesize{$(\pm 0.26)$}	& 30.52~\footnotesize{$(\pm 0.78)$}	& 49.58~\footnotesize{$(\pm 1.74)$}	& 16.33~\footnotesize{$(\pm 0.30)$}\\
        % & FC-LSTM~(\textbf{Ours})   & 21.61 & 34.21 & 14.81\% & 23.76 & 37.54  & 16.16\% & 27.99 & 44.01 & 19.11\% \\
         % Add other rows of METR-LA dataset
& STGODE & 21.88~\footnotesize{$(\pm 0.32)$}	& 33.70~\footnotesize{$(\pm 1.28)$}	& 14.72~\footnotesize{$(\pm 0.68)$}	& 24.71~\footnotesize{$(\pm 0.92)$}	& 39.08~\footnotesize{$(\pm 2.58)$}	& 15.94~\footnotesize{$(\pm 0.19)$}	& 30.00~\footnotesize{$(\pm 2.10)$}	& 48.24~\footnotesize{$(\pm 5.23)$}	& 18.62~\footnotesize{$(\pm 0.55)$}\\
& STGCN & 34.33~\footnotesize{$(\pm 0.82)$}	& 55.50~\footnotesize{$(\pm 1.37)$}	& 22.82~\footnotesize{$(\pm 2.84)$}	& 36.89~\footnotesize{$(\pm 0.82)$}	& 58.27~\footnotesize{$(\pm 1.29)$}	& 25.33~\footnotesize{$(\pm 3.06)$}	& 41.67~\footnotesize{$(\pm 1.25)$}	& 63.48~\footnotesize{$(\pm 1.37)$}	& 33.48~\footnotesize{$(\pm 3.80)$}\\
\cmidrule{2-11}
& TransGTR & \textbf{14.72}~\footnotesize{$(\pm 0.19)$}	& \textbf{22.87}~\footnotesize{$(\pm 0.27)$}	& \textbf{9.32~}\footnotesize{$(\pm 0.52)$}	& \textbf{15.89}~\footnotesize{$(\pm 0.31)$}	& \textbf{24.92}~\footnotesize{$(\pm 0.42)$}	& \textbf{10.00}~\footnotesize{$(\pm 0.69)$}	& \textbf{18.00}~\footnotesize{$(\pm 0.61)$}	& \textbf{28.32}~\footnotesize{$(\pm 0.79)$}	& \textbf{11.30}~\footnotesize{$(\pm 0.99)$}\\
 \cmidrule{2-11}
& \textbf{ST-FiT}& \underline{19.51}~\footnotesize{$(\pm 0.55)$}	& \underline{30.61}~\footnotesize{$(\pm 0.88)$}	& \underline{11.63}~\footnotesize{$(\pm 0.59)$}	& \underline{21.46}~\footnotesize{$(\pm 0.29)$}	& \underline{33.83}~\footnotesize{$(\pm 0.52)$}	& \underline{12.63}~\footnotesize{$(\pm 0.51)$}	& \underline{25.09}~\footnotesize{$(\pm 0.18)$}	& \underline{39.52}~\footnotesize{$(\pm 0.46)$}	& \underline{14.48}~\footnotesize{$(\pm 0.47)$}\\
 % Space before the next dataset
 
\bottomrule
\end{tabular}
}
\end{table*}

\subsection{Baseline Details}
Since the studied setting is novel, which requires the model to be inductive, we compare our framework to state-of-the-art baselines that can be adopted in such experimental settings. \textbf{Linear Sum:} \textit{(1)~Historical Average~(HA)}~\cite{dai2020hybrid} uses weighted averages from previous time steps as predictions for future periods. 
\noindent\textbf{Temporal-based}: \textit{(2) FC-LSTM}~\cite{sutskever2014sequence}. Long Short-Term Memory model with fully connected hidden units utilizes gating units to selectively capture temporal dependencies. \textbf{Spatial-Temporal}: \textit{(3) STGCN}~\cite{yu2017spatio} adopts a sandwich structure composed of the gated convolution network and ChebGCN to capture spatial-temporal dependencies. \textit{(4) STGODE}~\cite{fang2021spatial} learns both global semantic and spatial connections among nodes and captures spatial-temporal dynamics through a tensor-based ordinary differential equation (ODE). \textbf{Spatial-Temporal and Fine-tuning:} \textit{(5) TransGTR}~\cite{jin2023transferable} jointly learns and transfers graph structures and forecasting models across cities with knowledge distillation.

\subsection{Evaluation Metrics} We compare our method and baselines based on three commonly used metrics in time series forecasting: (1) Mean Absolute Error~(MAE) is the most used metric which reflects the prediction accuracy, (2) Root Mean Squared Error~(RMSE) is more sensitive to abnormal values, which could evaluate the reliability of models, (3) Mean Absolute Percentage Error~(MAPE) is scale-independent, which thus eliminates the influence of data units.~\cite{bandara2021improving}  The formulas of the three metrics are as follows:

\begin{equation}
    \begin{aligned}
& \operatorname{MAE}(x, \hat{x})=\frac{1}{|\Omega|} \sum_{i \in \Omega}\left|x_i-\hat{x}_i\right| \\
& \operatorname{RMSE}(x, \hat{x})=\sqrt{\frac{1}{|\Omega|} \sum_{i \in \Omega}\left(x_i-\hat{x}_i\right)^2} \\
& \operatorname{MAPE}(x, \hat{x})=\frac{1}{|\Omega|} \sum_{i \in \Omega} \frac{\left|x_i-\hat{x}_i\right|}{x_i},
\end{aligned}
\end{equation}
where $x_i$ denotes the $i$-th ground-truth, $\hat x_i$ denotes the $i$-th predicted values, and $\Omega$ is the indices of observed points.

\subsection{Implementation of ST-FiT}
We adopt a learning rate of 2e-2 and weight decay of 1e-3 for most experiments, together with commonly used number of epochs, e.g., 100~\cite{lan2022dstagnn}. We implement early stopping with patience parameter
10, where training stops if 10 epochs have passed without improvement of MAEs on the validation set. Unless otherwise stated, we set dimensions for all hidden representations to 64 for the temporal data augmentation module, the spatial topology learning module, and the STGNN backbone. We implement ST-FiT with Pytorch 2.0.1~\cite{paszke2017automatic} on multiple NVIDIA A6000 GPUs. The hyperparameters $\lambda$, $\epsilon$, and $\phi$ are set to 0.5, 0.9, and 0.1 respectively. The batch size is set to be 16, and we train the model with Adam optimizer. For the implementation of backbone model STGCN~\cite{yu2017spatio}, we follow settings of the implementation in \textbf{BasicTS}\footnotemark[1]\footnotetext[1]{https://github.com/zezhishao/BasicTS} except for the layer normalization which is not compatible with inductive forecasting task~\cite{liang2022basicts, shao2023exploring}.

\subsection{Implementation of Baselines}
For the implementation of all baselines, we adapt them to our inductive forecasting task by randomly sampling 10\% nodes for training. We set the random seed for selecting training nodes the same for all baselines, where the selecting process involves a BFS traversal on the original graph to preserve some connections between these training nodes. For the remaining implementation of \textbf{Historical Average}\footnotemark[1], \textbf{LSTM}\footnotemark[1], \textbf{STGCN}\footnotemark[2]\footnotetext[2]{https://github.com/hazdzz/STGCN}, \textbf{STGODE}\footnotemark[3]\footnotetext[3]{https://github.com/square-coder/STGODE}, and \textbf{TransGTR}\footnotemark[4]\footnotetext[4]{https://github.com/KL4805/TransGTR/}, we follow the same settings with those in their open-source code. 

% For the implementation of \textbf{TransGTR}, we adopt its official open-source code\footnote{https://github.com/KL4805/TransGTR/} for experiments.

\subsection{Packages Required for Implementations.}
We perform the experiments on a server with multiple Nvidia A6000 GPUs. Below we list the key packages and their associated versions in our implementation.
\begin{itemize}
    \item Python == 3.9.19
    \item torch == 2.2.2+cu121
    \item torch-geometric == 2.5.2
    \item cuda == 12.2
    \item numpy == 1.26.4
    \item pandas == 2.2.2
    \item scikit-learn == 1.4.2
    \item pandas == 2.2.2
    \item scipy == 1.13.0
\end{itemize}

\section{Supplementary Experiments}

\subsection{Generalization Performance}
In this subsection, we present additional experimental results to answer RQ1. Specifically, we compare the average performance of the first 3, 6, 12 time steps in the test window of the temporal data of forecasting on the MAE, RMSE, and MAPE metrics, where we refer to such time steps as \textbf{Horizons}. We make observation from Table~\ref{tab:main_full} as follows. (1) ST-FiT outperforms all baselines  that do not require fine-tuning. This verifies the effectiveness of ST-FiT in generalizing to the nodes with different temporal dependencies. (2) ST-FiT achieves comparable performance with fine-tuned model TransGTR, which corroborates the effectiveness of ST-FiT in capturing diverse spatial-temporal dependencies.

\begin{table*}[!htbp]
\vspace{4mm}
\setlength{\tabcolsep}{3.65pt}
\renewcommand{\arraystretch}{1.15}
\centering
\caption{Performance comparison on 3, 6, 12 horizons of forecasting for ablation study. The best results are in bold, and the second best results are underlined. It is observed that removing any module of ST-FiT will jeopardize the overall performance.}
\label{tab:ablation_full}
\resizebox{\textwidth}{!}{
\begin{tabular}{
  l % Dataset column
  l % Methods column
ccc % Horizon 3 columns
  ccc % Horizon 6 columns
  ccc % Horizon 12 columns
}
\toprule
Datasets & {Variants} & \multicolumn{3}{c}{Horizon 3} & \multicolumn{3}{c}{Horizon 6} & \multicolumn{3}{c}{Horizon 12} \\
\cmidrule(lr){3-5}
\cmidrule(lr){6-8}
\cmidrule(lr){9-11}
 & & {MAE} & {RMSE} & {MAPE (\%)} & {MAE} & {RMSE} & {MAPE (\%)} & {MAE} & {RMSE} & {MAPE (\%)} \\
\midrule
\multirow{7}{*}{PEMS03} & \textbf{ST-FiT} & \underline{14.71}~\footnotesize{$(\pm 0.30)$}	& \underline{22.80}~\footnotesize{$(\pm 0.34)$}	& 16.53~\footnotesize{$(\pm 1.16)$}	& \textbf{16.03}~\footnotesize{$(\pm 0.25)$}	& \textbf{25.21}~\footnotesize{$(\pm 0.36)$}	& 17.40~\footnotesize{$(\pm 1.60)$}	& \textbf{18.40}~\footnotesize{$(\pm 0.23)$}	& \textbf{29.31}~\footnotesize{$(\pm 0.32)$}	& \underline{18.94}~\footnotesize{$(\pm 1.53)$}\\
 \cmidrule{2-11}
 & \textit{w/o aug}& 15.31~\footnotesize{$(\pm 1.07)$}	& 23.96~\footnotesize{$(\pm 1.64)$}	& 16.82~\footnotesize{$(\pm 4.12)$}	& 16.73~\footnotesize{$(\pm 1.03)$}	& 26.49~\footnotesize{$(\pm 1.64)$}	& 17.59~\footnotesize{$(\pm 3.78)$}	& 19.44~\footnotesize{$(\pm 1.28)$}	& 30.94~\footnotesize{$(\pm 1.92)$}	& 19.07~\footnotesize{$(\pm 3.22)$}\\
 & \textit{w/o sim}&  \textbf{14.70}~\footnotesize{$(\pm 0.03)$}	& \textbf{22.56}~\footnotesize{$(\pm 0.14)$}	& 18.43~\footnotesize{$(\pm 6.40)$}	& 16.21~\footnotesize{$(\pm 0.21)$}	& \underline{25.22}~\footnotesize{$(\pm 0.54)$}	& 19.69~\footnotesize{$(\pm 6.59)$}	& 18.83~\footnotesize{$(\pm 0.59)$}	& 29.75~\footnotesize{$(\pm 1.10)$}	& 21.78~\footnotesize{$(\pm 5.60)$}\\

 & \textit{w/o fst}& 15.32~\footnotesize{$(\pm 0.05)$}	& 23.62~\footnotesize{$(\pm 0.30)$}	& 19.82~\footnotesize{$(\pm 3.23)$}	& 16.76~\footnotesize{$(\pm 0.11)$}	& 25.94~\footnotesize{$(\pm 0.20)$}	& 21.84~\footnotesize{$(\pm 3.99)$}	& 19.58~\footnotesize{$(\pm 0.37)$}	& 30.30~\footnotesize{$(\pm 0.06)$}	& 26.14~\footnotesize{$(\pm 6.58)$}\\
          \cmidrule{2-11}
 & \textit{w/o gl}& 15.25~\footnotesize{$(\pm 0.08)$}	& 23.96~\footnotesize{$(\pm 0.26)$}	& 16.09~\footnotesize{$(\pm 0.08)$}	& 17.11~\footnotesize{$(\pm 0.02)$}	& 27.10~\footnotesize{$(\pm 0.09)$}	& 17.48~\footnotesize{$(\pm 0.18)$}	& 20.72~\footnotesize{$(\pm 0.03)$}	& 33.05~\footnotesize{$(\pm 0.05)$}	& 21.39~\footnotesize{$(\pm 0.25)$}\\
 & \textit{w/o gs}& \underline{14.71}~\footnotesize{$(\pm 0.19)$}	& 22.97~\footnotesize{$(\pm 0.42)$}	& \textbf{14.42}~\footnotesize{$(\pm 0.20)$}	& \underline{16.06}~\footnotesize{$(\pm 0.13)$}	& 25.39~\footnotesize{$(\pm 0.29)$}	& \textbf{15.70}~\footnotesize{$(\pm 0.47)$}	& \underline{18.48}~\footnotesize{$(\pm 0.03)$}	& \underline{29.58}~\footnotesize{$(\pm 0.08)$}	& \textbf{17.88}~\footnotesize{$(\pm 0.03)$}\\
 & \textit{identity}& 15.16~\footnotesize{$(\pm 0.03)$}	& 23.38~\footnotesize{$(\pm 0.01)$}	& \underline{15.64}~\footnotesize{$(\pm 1.63)$}	& 17.05~\footnotesize{$(\pm 0.03)$}	& 26.69~\footnotesize{$(\pm 0.02)$}	& \underline{17.17}~\footnotesize{$(\pm 1.30)$}	& 20.65~\footnotesize{$(\pm 0.01)$}	& 32.72~\footnotesize{$(\pm 0.11)$}	& 20.75~\footnotesize{$(\pm 1.32)$}\\
 \cmidrule{1-11}
\multirow{7}{*}{PEMS04} & \textbf{ST-FiT} & \textbf{20.31}~\footnotesize{$(\pm 0.34)$}	& \textbf{31.97}~\footnotesize{$(\pm 0.54)$}	& \textbf{14.06}~\footnotesize{$(\pm 0.10)$}	& \textbf{21.95}~\footnotesize{$(\pm 0.35)$}	& \textbf{34.48}~\footnotesize{$(\pm 0.55)$}	& \textbf{15.13}~\footnotesize{$(\pm 0.26)$}	& \textbf{25.11}~\footnotesize{$(\pm 0.42)$}	& \textbf{39.30}~\footnotesize{$(\pm 0.62)$}	& \textbf{17.23}~\footnotesize{$(\pm 0.43)$}\\
 \cmidrule{2-11}
 & \textit{w/o aug}& 21.40~\footnotesize{$(\pm 0.35)$}	& 33.95~\footnotesize{$(\pm 0.60)$}	& 14.55~\footnotesize{$(\pm 0.64)$}	& 23.36~\footnotesize{$(\pm 0.30)$}	& 37.00~\footnotesize{$(\pm 0.45)$}	& 15.77~\footnotesize{$(\pm 0.47)$}	& 27.31~\footnotesize{$(\pm 0.18)$}	& 43.04~\footnotesize{$(\pm 0.29)$}	& 18.64~\footnotesize{$(\pm 0.10)$}\\
 & \textit{w/o sim}& \underline{20.61}~\footnotesize{$(\pm 0.02)$}	& \underline{32.54}~\footnotesize{$(\pm 0.11)$}	& \underline{14.19}~\footnotesize{$(\pm 0.23)$}	& \underline{22.26}~\footnotesize{$(\pm 0.01)$}	& \underline{35.14}~\footnotesize{$(\pm 0.17)$}	& \underline{15.07}~\footnotesize{$(\pm 0.20)$}	& \underline{25.55}~\footnotesize{$(\pm 0.14)$}	& \underline{40.18}~\footnotesize{$(\pm 0.44)$}	& \underline{17.41}~\footnotesize{$(\pm 0.16)$}\\

 & \textit{w/o fst}& 21.35~\footnotesize{$(\pm 0.76)$}	& 33.59~\footnotesize{$(\pm 1.39)$}	& 15.79~\footnotesize{$(\pm 0.23)$}	& 23.10~\footnotesize{$(\pm 0.87)$}	& 36.36~\footnotesize{$(\pm 1.46)$}	& 16.59~\footnotesize{$(\pm 0.13)$}	& 26.61~\footnotesize{$(\pm 1.06)$}	& 41.66~\footnotesize{$(\pm 1.61)$}	& 19.00~\footnotesize{$(\pm 0.56)$}\\
          \cmidrule{2-11}
 & \textit{w/o gl}& 21.24~\footnotesize{$(\pm 0.28)$}	& 33.77~\footnotesize{$(\pm 0.50)$}	& 14.56~\footnotesize{$(\pm 0.29)$}	& 23.37~\footnotesize{$(\pm 0.29)$}	& 37.07~\footnotesize{$(\pm 0.57)$}	& 15.94~\footnotesize{$(\pm 0.20)$}	& 27.58~\footnotesize{$(\pm 0.30)$}	& 43.50~\footnotesize{$(\pm 0.55)$}	& 19.00~\footnotesize{$(\pm 0.28)$}   \\
 & \textit{w/o gs}& 21.33~\footnotesize{$(\pm 0.76)$}	& 33.99~\footnotesize{$(\pm 1.63)$}	& 14.55~\footnotesize{$(\pm 0.35)$}	& 23.20~\footnotesize{$(\pm 0.95)$}	& 37.00~\footnotesize{$(\pm 1.98)$}	& 15.58~\footnotesize{$(\pm 0.41)$}	& 26.68~\footnotesize{$(\pm 1.12)$}	& 42.55~\footnotesize{$(\pm 2.20)$}	& 17.76~\footnotesize{$(\pm 0.35)$}\\
 & \textit{identity}& 21.58~\footnotesize{$(\pm 0.31)$}	& 34.76~\footnotesize{$(\pm 0.75)$}	& 14.56~\footnotesize{$(\pm 0.42)$}	& 23.78~\footnotesize{$(\pm 0.24)$}	& 38.15~\footnotesize{$(\pm 0.54)$}	& 16.09~\footnotesize{$(\pm 0.60)$}	& 28.00~\footnotesize{$(\pm 0.18)$}	& 44.48~\footnotesize{$(\pm 0.36)$}	& 19.38~\footnotesize{$(\pm 0.60)$}\\
\cmidrule{1-11}
 \multirow{7}{*}{PEMS08} & \textbf{ST-FiT}  & \textbf{19.51}~\footnotesize{$(\pm 0.55)$}	& \textbf{30.61}~\footnotesize{$(\pm 0.88)$}	& \textbf{11.63}~\footnotesize{$(\pm 0.59)$}	& \textbf{21.46}~\footnotesize{$(\pm 0.29)$}	& \textbf{33.83}~\footnotesize{$(\pm 0.52)$}	& \textbf{12.63}~\footnotesize{$(\pm 0.51)$}	& \textbf{25.09}~\footnotesize{$(\pm 0.18)$}	& \textbf{39.52}~\footnotesize{$(\pm 0.46)$}	& \textbf{14.48}~\footnotesize{$(\pm 0.47)$}\\
 \cmidrule{2-11}
 & \textit{w/o aug}& 21.34~\footnotesize{$(\pm 0.59)$}	& 34.04~\footnotesize{$(\pm 1.43)$}	& 12.68~\footnotesize{$(\pm 0.28)$}	& 23.54~\footnotesize{$(\pm 0.81)$}	& 37.53~\footnotesize{$(\pm 1.69)$}	& 13.83~\footnotesize{$(\pm 0.09)$}	& 27.69~\footnotesize{$(\pm 1.00)$}	& 43.99~\footnotesize{$(\pm 1.77)$}	& 16.11~\footnotesize{$(\pm 0.19)$}\\
 & \textit{w/o sim}& \underline{20.49}~\footnotesize{$(\pm 2.32)$}	& \underline{31.84}~\footnotesize{$(\pm 3.97)$}	& 12.13~\footnotesize{$(\pm 1.16)$}	& \underline{22.42}~\footnotesize{$(\pm 2.12)$}	& \underline{35.10}~\footnotesize{$(\pm 3.52)$}	& \underline{13.12}~\footnotesize{$(\pm 1.20)$}	& \underline{26.35}~\footnotesize{$(\pm 2.10)$}	& \underline{41.43}~\footnotesize{$(\pm 3.37)$}	& \underline{15.07}~\footnotesize{$(\pm 1.22)$}\\

 & \textit{w/o fst}& 22.54~\footnotesize{$(\pm 0.13)$}	& 35.51~\footnotesize{$(\pm 1.03)$}	& 13.56~\footnotesize{$(\pm 0.10)$}	& 24.77~\footnotesize{$(\pm 0.43)$}	& 38.92~\footnotesize{$(\pm 0.30)$}	& 14.35~\footnotesize{$(\pm 0.06)$}	& 28.97~\footnotesize{$(\pm 1.05)$}	& 45.15~\footnotesize{$(\pm 0.88)$}	& 16.31~\footnotesize{$(\pm 0.04)$}\\
          \cmidrule{2-11}
 & \textit{w/o gl}& 21.68~\footnotesize{$(\pm 1.08)$}	& 35.83~\footnotesize{$(\pm 1.77)$}	& \underline{11.93}~\footnotesize{$(\pm 0.29)$}	& 24.02~\footnotesize{$(\pm 1.12)$}	& 39.55~\footnotesize{$(\pm 1.53)$}	& 13.17~\footnotesize{$(\pm 0.30)$}	& 28.16~\footnotesize{$(\pm 1.46)$}	& 46.13~\footnotesize{$(\pm 2.36)$}	& 15.59~\footnotesize{$(\pm 0.42)$}\\
 & \textit{w/o gs}& 25.60~\footnotesize{$(\pm 3.24)$}	& 38.52~\footnotesize{$(\pm 4.45)$}	& 14.03~\footnotesize{$(\pm 1.21)$}	& 29.05~\footnotesize{$(\pm 3.05)$}	& 43.83~\footnotesize{$(\pm 3.87)$}	& 15.19~\footnotesize{$(\pm 1.32)$}	& 35.03~\footnotesize{$(\pm 2.44)$}	& 52.67~\footnotesize{$(\pm 2.42)$}	& 17.51~\footnotesize{$(\pm 1.30)$}\\
 & \textit{identity}& 21.41~\footnotesize{$(\pm 1.28)$}	& 34.54~\footnotesize{$(\pm 2.67)$}	& 11.99~\footnotesize{$(\pm 0.42)$}	& 23.97~\footnotesize{$(\pm 1.33)$}	& 38.93~\footnotesize{$(\pm 2.66)$}	& 13.27~\footnotesize{$(\pm 0.34)$}	& 28.35~\footnotesize{$(\pm 1.44)$}	& 45.97~\footnotesize{$(\pm 2.67)$}	& 15.57~\footnotesize{$(\pm 0.40)$}\\
% \addlinespace % Space before the next dataset

% \addlinespace % Space before the next dataset
 
\bottomrule
\end{tabular}
}
\end{table*}

\subsection{Performance w.r.t. Training Node Ratios}
In this subsection, we present additional results for evaluation of model performance under different limitation levels of training data. We present all possible combinations of datasets and metrics, which results in 9 groups of experiments in total. From Figure~\ref{fig:2}, we could observe the following key findings: (1) ST-FiT consistently outperforms all other baselines almost under all ratios of training nodes in inductive forecasting task, which further corroborates the effectiveness of ST-FiT for the inductive forecasting task. This demonstrates its reliability with varying amounts of training data. (2) ST-FiT could achieve competitive performance where training nodes ratio reachs up to 100\% (no inductive forecasting), which demonstrate ST-FiT's practicality for both inductive and non-inductive setting.

\subsection{Ablation Study}
In this subsection, we provide additional experimental results to answer RQ3. The results are shown in Table~\ref{tab:ablation_full}. We make the observations as follows: (1) Both the temporal data augmentation and spatial topology learning modules effectively contribute to the overall performance, which verifies both of their effectiveness for improving generalization. (2) Removing either $\mathcal{L}_{sim}$ or $\mathcal{L}_{fst}$ degrades the performance, which indicates their ability in generating temporal data with diverse temporal dependencies for training. (3) Spatial topology learning based on Gumbel-Softmax performs better than all its variants, which verifies the effectiveness of spatial topology learning. Meanwhile, ST-FiT outperforms all other variants, which indicates that the learned spatial topology significantly enhances the diversity of spatial dependencies and thus benefits the forecasting.

\begin{figure}[!t]
    \centering
    \includegraphics[width=\columnwidth]{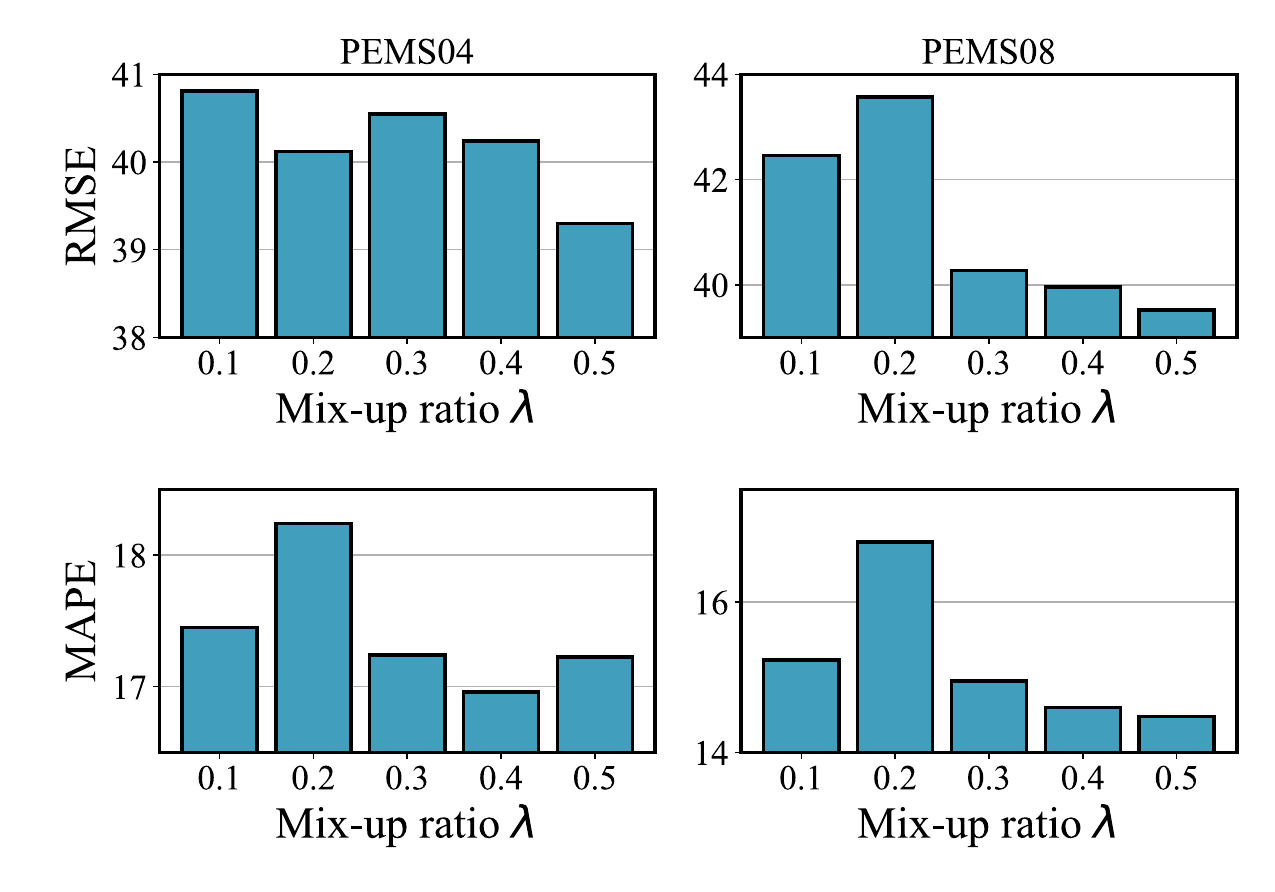}
    \caption{Performance of ST-FiT with different values of mix-up ratio $\lambda$. The mix-up ratio $\lambda$ exhibits slight influence on performance, while the positive correlation between mix-up ratio $\lambda$ and the performance still exhibits.}
    \label{fig:1}
\end{figure}

\begin{figure}[!h]
    \centering
    \includegraphics[width=\columnwidth]{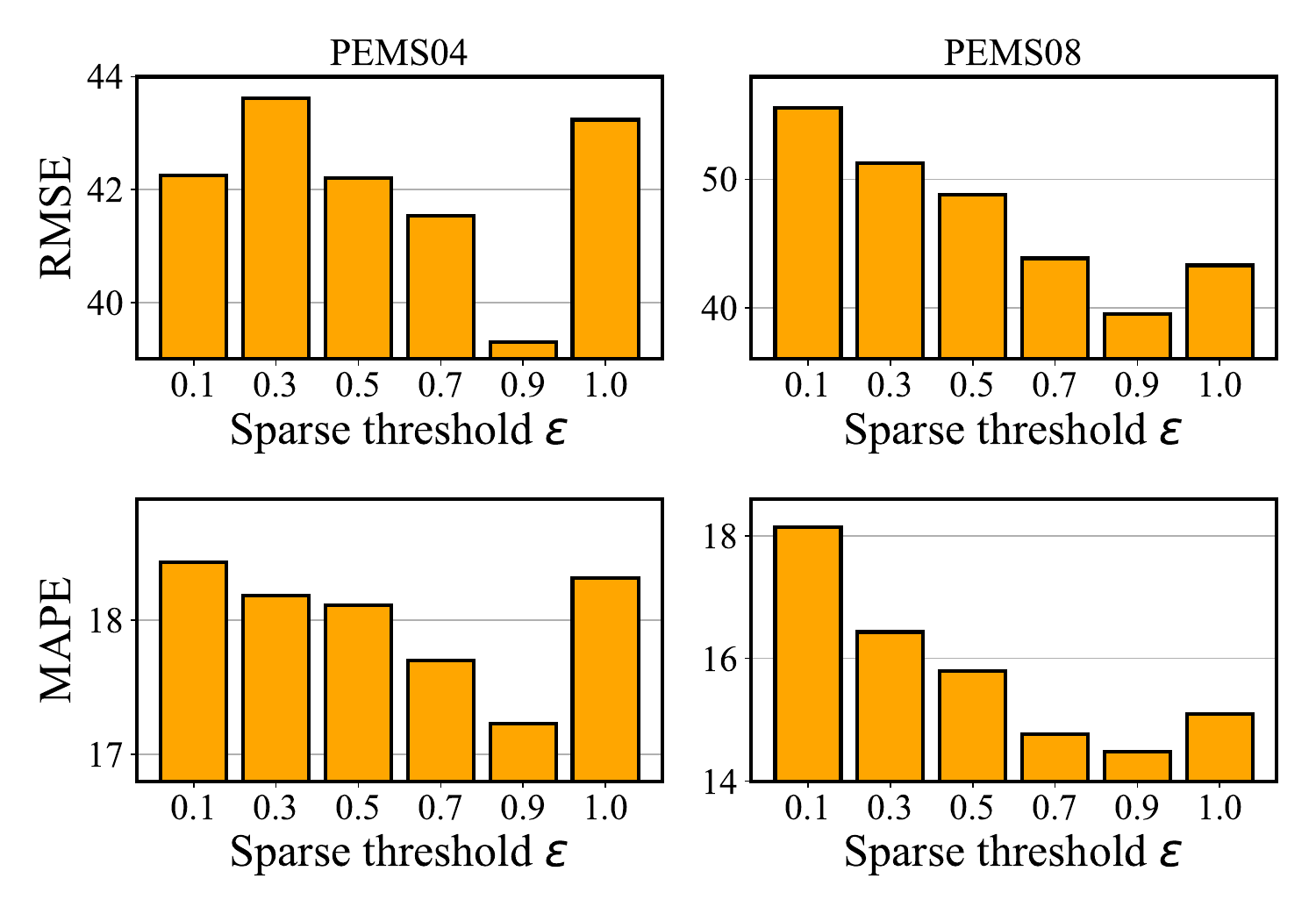}
    \caption{Performance of ST-FiT with different sparse thresholds $\o$. Sparsity has positive correlation with performance, while extreme level of sparsity might bring negative influence due to loss of several key connections.}
    \label{fig:3}
\end{figure}

\subsection{Parameter Sensitivity}
In this subsection, we present additional experimental results for different choices of mix-up ratio $\lambda$ and sparse threshold $\epsilon$. Here we provide RMSE and MAPE results for PEMS04 and PEMS08. From Figure~\ref{fig:1}, we could observe that (1) mix-up ratio has slight impact on performances, which can result from the sufficient temporal dependencies acquired from even a small mix-up ratio such as $10\%$. (2) Higher mix-up ratio exhibits better performances in most cases, which owes to their introduction of more diverse temporal dependencies. From Figure~\ref{fig:3}, we could observe that (1) performance achieves the best when sparse threshold $\epsilon$ is 0.9, which demonstrate the effectiveness of adopting sparse structure. (2) performance decreases when sparse threshold further increases, which can result from the loss of core spatial dependencies. From the results above, we recommended setting $\lambda$ to 0.5 and $\epsilon$ to 0.9.

\subsection{Efficiency Study}

\begin{figure}[!h]
    \centering
    \includegraphics[width=0.95\columnwidth]{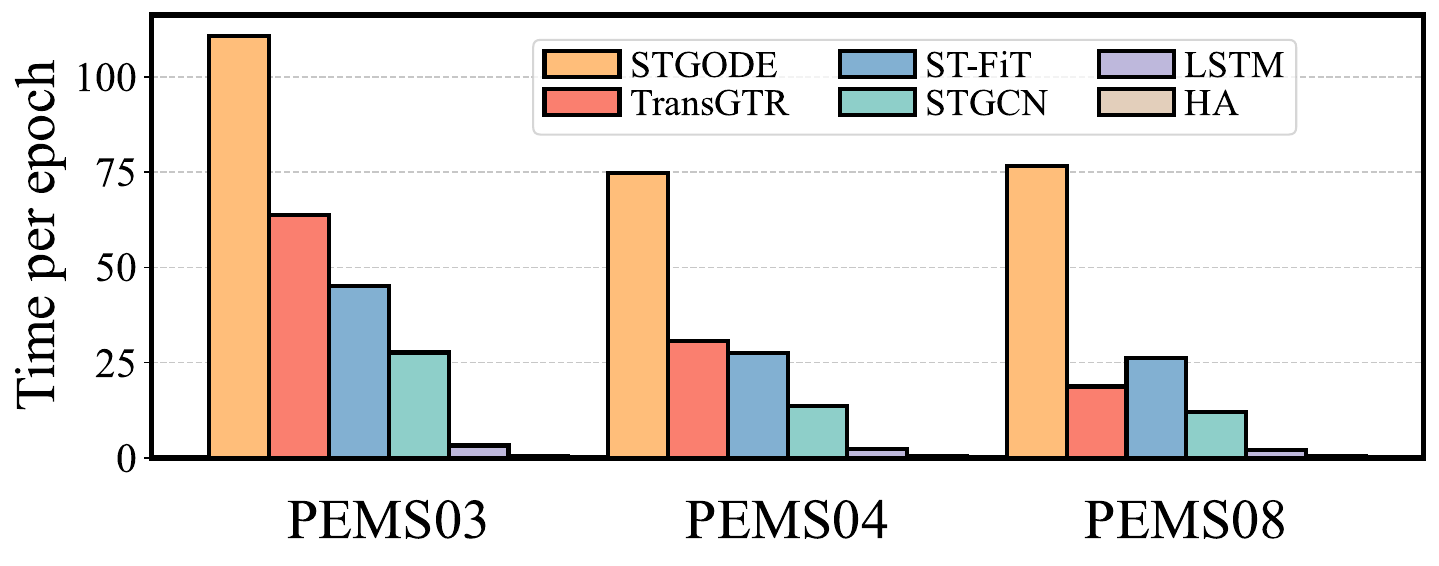}
    \caption{Comparison of training time per epoch between ST-FiT and different types of baseline models. ST-FiT demonstrates competitive or even superior efficiency compared to fine-tuning baseline TransGTR (pre-training).}
    \label{fig:efficiency}
\end{figure}

In this subsection, we analyse the efficiency of ST-FiT compared to baselines. We present the average time consumed per each epoch in Table~\ref{fig:efficiency} (a complete iteration for ST-FiT), from which we make the following observations: (1) The time consumed per epoch by ST-FiT is nearly double that of STGCN, which indicates that the primary reason for the increased time in ST-FiT is the two forward passes per full iteration, whereas temporal data augmentation and spatial topology learning have small impact on the time consumption. (2)The efficiency of ST-FiT is competitive to TransGTR during the pre-training stage. Given that TransGTR includes additional distilling and fine-tuning processes, it struggles to satisfy the efficiency requirements of some real world scenarios.

% $ and sparse threshold $\epsilon$. Here we provide RMSE and MAPE results for PEMS04 and PEMS08. From Figure~\ref{fig:1}, we could observe that (1) mix-up ratio has slight impact on performances, which can result from the sufficient temporal dependencies acquired from even a small mix-up ratio such as $10\%$. (2) Higher mix-up ratio exhibits better performances in most cases, which owes to their introduction of more diverse temporal dependencies. From Figure~\ref{fig:3}, we could observe that (1) performance achieves the best when sparse threshold $\epsilon$ is 0.9, which demonstrate the effectiveness of adopting sparse structure. (2) performance decreases when sparse threshold further increases, which can result from the loss of core spatial dependencies. From the results above, we recommended setting $\lambda$ to 0.5 and $\epsilon$ to 0.9.

\section{Algorithm}
In this section, we present the complete algorithmic routine of ST-FiT in Algorithm~\ref{alg:cap}.
\begin{algorithm}
\caption{Training and Inference of ST-FiT}\label{alg:cap}
\begin{algorithmic}[1]
\State \textbf{Training Phase:}
\State \textbf{Input:} $\{\boldsymbol{X}_{train}^{t-\kappa:t}\}$:  Features of $\kappa$ time steps from nodes for training; $\{\boldsymbol{X}_{train}^{t+1:t+\tau}\}$: Features in following $\tau$ time steps from nodes for training; $\boldsymbol{A}$: Adjacency matrix; $\eta$: Hyper-parameters in objectives; $ \epsilon$: Threshold for sparse topology learning; $\phi$: Temperature in spatial topology learning; $s $: Temperature for Gumbel-Softmax reparameterization;

\If{$A$ is None}
    \State $\boldsymbol{A} \leftarrow $ Initialized from cosine similarity of hidden representations which are randomly initialized
    % Cosine \ similarity \ of \ random \ initialized \ node \ features
\EndIf

% \State 
\State $\boldsymbol{\theta}^{aug}, \boldsymbol{\theta}^{gf} \leftarrow Random\ Initialize$
\While{Stopping condition is not met}
\State \# $Phase 1$
\State Generate temporal data following Eq. (4) in the main body;
% $\{\boldsymbol{\hat{X}}^{t-\kappa:t}\}$ = Augmentation($\{\boldsymbol{X}_{train}^{t-\kappa:t}\}$)
\State Learn spatial topology following Eq. (7) in the main body;
% \State $\boldsymbol{A}^l$ = $\operatorname{LearnGraph}(\{\boldsymbol{X}_{train}^{t-\kappa:t}\}, \{\boldsymbol{\hat{X}}^{t-\kappa:t}\}, \boldsymbol{A})$
\State Compute $\mathcal{L}_{aug}$ following Eq. (10) in the main body;
% \State $\mathcal{L}_{aug} = \operatorname{ComputeAugLoss}(f(\{\boldsymbol{X}_{train}^{t-\kappa:t}\},$
% \State \ \ \ \ \ \ \ \ \ \ \ \ \ $\{\boldsymbol{\hat{X}}^{t-\kappa:t}\}, \boldsymbol{A}^l) \ , \{\boldsymbol{X}_{train}^{t+1:t+\tau}\}$
\State Update the weights of $f$ with gradient-based techniques;
\State 
% \State $\theta^{aug} \leftarrow \theta^{aug} - \eta\cdot \nabla_{\theta^{aug}}\mathcal{L}_{aug}$
\State \# $Phase 2$
\State Generate temporal data following Eq. (4) in the main body;
\State Learn spatial topology following Eq. (7) in the main body;
\State Compute $\mathcal{L}_{aug}$ following Eq. (12) in the main body;
\State Update the weights of $f$ with gradient-based techniques;
% \State $\{\boldsymbol{\hat{X}}^{t-\kappa:t}\}$ = Augmentation($\{\boldsymbol{X}_{train}^{t-\kappa:t}\}$)
% \State $\boldsymbol{A}^l$ = $\operatorname{LearnGraph}(\{\boldsymbol{X}_{train}^{t-\kappa:t}\}, \{\boldsymbol{\hat{X}}^{t-\kappa:t}\}, \boldsymbol{A})$
% \State $\mathcal{L}_{gf} = \operatorname{ComputeForecastLoss}(f(\{\boldsymbol{X}_{train}^{t-\kappa:t}\}, $
% \State  \ \ \ \ \ \ \ \ \ \ \ $\{\boldsymbol{\hat{X}}^{t-\kappa:t}\}, \boldsymbol{A}^l) \ , \{\boldsymbol{X}_{train}^{t+1:t+\tau}\}$
% \State $\theta^{gf} \leftarrow \theta^{gf} - \eta\cdot \nabla_{\theta^{gf}}\mathcal{L}_{gf}$
\EndWhile
\State
\State \textbf{Inference Phase:}
\State \textbf{Input:} $\{\boldsymbol{X}^{t-\kappa:t}\}$: Features of $\kappa$ time steps; $ \epsilon$: Threshold for sparse topology learning; $\phi$: Temperature in spatial topology learning; $s $: Temperature for Gumbel-Softmax;
\State Generate spatial topology following Eq. (7) in the main body;
% \State $\boldsymbol{A}^l$ = $\operatorname{LearnGraph}(\{\boldsymbol{X}^{t-\kappa:t}\})$
\State Conduct forecasting following Eq. (1) in the main body;
% \State $\{\boldsymbol{X}^{t-1:t+\tau}\} = f(\{\boldsymbol{X}^{t-\kappa:t}\}, \boldsymbol{A}^l)$
\State \textbf{return} $\{\boldsymbol{X}^{t-1:t+\tau}\}$
\end{algorithmic}
\end{algorithm}

% \bibliography{aaai24}

\end{document}